\documentclass{article}

\PassOptionsToPackage{numbers, compress}{natbib}

\usepackage[final]{neurips_2022}

\usepackage[pagebackref=true,breaklinks=true,citecolor=tealblue,colorlinks,bookmarks=false]{hyperref}

\usepackage[utf8]{inputenc} 
\usepackage[T1]{fontenc}    
\usepackage{url}            
\usepackage{booktabs}       
\usepackage{amsfonts}       
\usepackage{nicefrac}       
\usepackage{microtype}      
\usepackage{colortbl}         
\usepackage{multirow}
\usepackage{tabularx}
\usepackage{graphicx}
\usepackage{amsmath}
\usepackage{amssymb}
\usepackage{makecell}
\usepackage[dvipsnames, table, x11names]{xcolor}
\usepackage{wrapfig}
\usepackage{setspace}

\newcommand{\ie}{\textit{i.e. }}
\newcommand{\eg}{\textit{e.g. }}
\newcommand{\vs}{\textit{vs. }}
\newcommand{\etal}{\textit{et al. }}

\title{Towards Efficient 3D Object Detection with Knowledge Distillation}

\author{%
  Jihan Yang\textsuperscript{1} \quad
  Shaoshuai Shi\textsuperscript{2} \quad
  Runyu Ding\textsuperscript{1} \quad
  Zhe Wang\textsuperscript{3} \quad
  Xiaojuan Qi\textsuperscript{1} \vspace{.2em}\\
  \textsuperscript{1}The University of Hong Kong \quad \textsuperscript{2}Max Planck Institute for Informatics \quad \textsuperscript{3}SenseTime Research \\
  {\tt\small \{jhyang, ryding, xjqi\}@eee.hku.hk, \{shaoshuaics, wzlewis16\}@gmail.com}
}

\begin{document}
\definecolor{tealblue}{HTML}{0071bc}
\maketitle

\begin{abstract}
  Despite substantial progress in 3D object detection, advanced 3D detectors often suffer from heavy computation overheads. 
  To this end, we explore the potential of knowledge distillation (KD) for developing efficient 3D object detectors, focusing on popular pillar- and voxel-based detectors.
  In the absence of well-developed teacher-student pairs, we first study how to obtain student models with good trade offs between accuracy and efficiency from the perspectives of model compression and input resolution reduction. 
  Then, we build a benchmark to assess existing KD methods developed in the 2D domain for 3D object detection upon six well-constructed teacher-student pairs.
  Further, we propose an improved KD pipeline incorporating an enhanced logit KD method that performs KD on only a few pivotal positions determined by teacher classification response, and a teacher-guided student model initialization to facilitate transferring teacher model's feature extraction ability to students through weight inheritance.  
  Finally, we conduct extensive experiments on the Waymo and KITTI dataset. Our best performing model achieves $65.75\%$ LEVEL 2 mAPH, surpassing its teacher model and requiring only $44\%$ of teacher flops on Waymo. Our most efficient model runs 51 FPS on an NVIDIA A100, which is $2.2\times$ faster than PointPillar with even higher accuracy on Waymo. Code is available at \url{https://github.com/CVMI-Lab/SparseKD}.
 

\end{abstract}

\section{Introduction}
\begin{wrapfigure}{r}{6cm}
\vspace{-0.8cm}
\begin{center}
\includegraphics[scale=0.46]{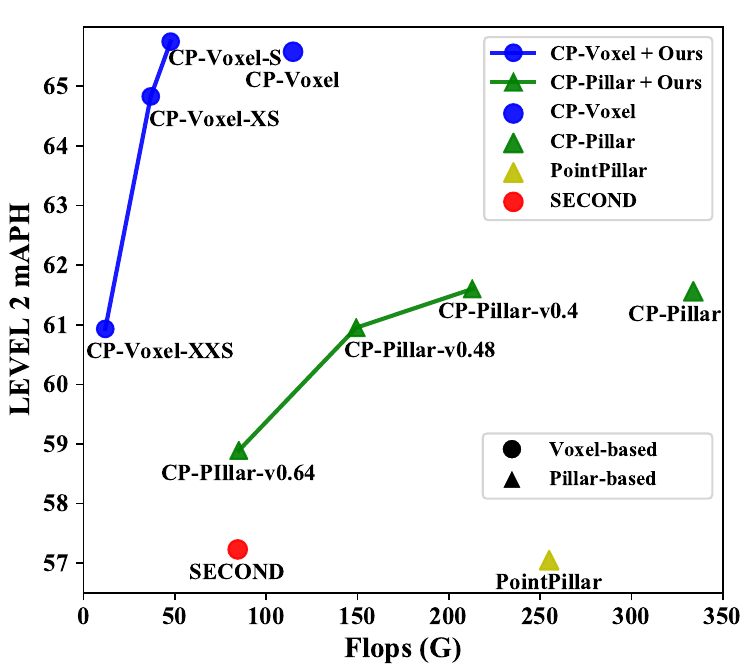}
\end{center}
\vspace{-0.6cm}
\caption{Performance and flops comparison of single-stage detectors on Waymo: CP-Pillar \cite{yin2021center}, CP-Voxel \cite{yin2021center}, SECOND \cite{yan2018second}, PointPillar~\cite{lang2019pointpillars} and Ours.}
\vspace{-0.5cm}
\label{fig:performance}
\end{wrapfigure}

\vspace{-0.2cm}
3D object detection from point clouds is a fundamental perception task with broad applications on autonomous driving, robotics and smart city, etc. Recently, benefited from large-scale 3D perception datasets~\cite{geiger2012we,caesar2020nuscenes,sun2020scalability} and advanced point- \cite{qi2017pointnet++}, pillar- \cite{lang2019pointpillars,wang2020pillar} and voxel-based \cite{graham2017submanifold,zhou2018voxelnet} representations of sparse and irregular LiDAR point cloud scenes, 3D detection has achieved remarkable progress~\cite{yan2018second,shi2019pointrcnn,shi2020pv,bewley2020range,yin2021center}. 
However, stronger performance is often accompanied with heavier computation burden (see Figure~\ref{fig:performance}), rendering their adoption in real-world applications still a challenging problem.


Recent attempts to improve efficiency focus on developing specified architectures for point-based 3D object detectors~\cite{chen2019fast,zhang2022not}, not generalizable to a wide spectrum of pillar/voxel-based methods~\cite{zhou2018voxelnet,lang2019pointpillars,yan2018second,shi2020pv,yin2021center,deng2020voxel}. Here, we aim at a model-agnostic framework for obtaining efficient and accurate 3D object detectors with knowledge distillation (KD).
Due to its effectiveness, generality and simplicity, KD has become a de facto strategy to develop efficient models in a variety of 2D tasks \cite{hinton2015distilling,liu2019structured,dai2021general,hou2019learning,xing2021categorical}. It facilitates improving the performance of a lightweight and efficient student model by harvesting knowledge learned by an accurate yet computationally heavy teacher model.
Despite of numerous studies in 2D tasks, the investigation of KD for efficient 3D object detection has largely escaped research attention with unresolved research challenges. In this paper, we conduct the first systematic study on knowledge distillation for developing high-performance and efficient 3D LiDAR-based detectors.

First, we study \textit{how to obtain lightweight student detectors with satisfactory efficiency and accuracy trade offs given a pre-trained teacher 3D object detector}. 
Different from the 2D domain where  well-developed backbone architectures with different model efficiencies are readily available (e.g., ResNet 18 \vs ResNet 50)~\cite{simonyan2014very,szegedy2015going,he2016deep}, such scalable backbones in 3D scenes are still under-explored. This makes the design of suitable student models a non-trivial problem. Intuitively, a good student model should achieve a good compromise between accuracy and efficiency as a poor student model may have inferior capabilities or architectural level shortcomings, causing difficulties for further knowledge distillation.
In this perspective, we first propose the Cost Performance Ratio (CPR) to fairly evaluate a model in terms of both efficiency and capability. 
Then, we study different factors including model width (\ie number of channels), depth {(\ie number of layers)} and input resolution on both pillar-based and voxel-based architectures.
Specifically, we find that pillar-based architecture favors input-level compression (\ie reduce input resolution) while voxel-based detector prefers width-level compression, due to less spatial redundancy of voxel-based detectors.

Second, we empirically investigate \textit{the effectiveness of existing  knowledge distillation methods on this new setting} upon accurate teacher models and efficient student models. As there is no prior research for this problem, we benchmark seven existing  knowledge distillation methods on top of six teacher-student pairs covering voxel- and pillar-based architectures. Specifically, we evaluate the following major streams: \textbf{logit KD} distilling on model {outputs} {(KD~\cite{hinton2015distilling} and GID-L~\cite{dai2021general})}, \textbf{feature KD} mimicking intermediate features {(FitNet~\cite{romero2014fitnets}, Mimic~\cite{li2017mimicking}, FG~\cite{wang2019distilling} and GID-F~\cite{dai2021general})} and \textbf{label KD} leveraging teacher's predictions for label assignment~\cite{nguyen2022improving}. Further, we also study their synergy effects and empirically find that feature KD outperforms others individually, but fails to cooperate with other KD manners to further boost the performance in 3D detection. 
This is potentially caused by the fact that logit and label KD can add an implicit regularization to the intermediate features. The best strategy that we obtain through empirical studies is to use FG~\cite{wang2019distilling} or Mimic~\cite{li2017mimicking} individually.

Third, we propose \textit{simple, general, and effective strategies to improve knowledge distillation on 3D object detection} upon the strong KD baseline derived as above. Motivated by the extreme imbalance between {small informative areas containing 3D objects} and large redundant background areas in 3D scenes, 
we design a modified logit KD method, namely pivotal position logit KD, 
enforcing imitation on only locations with highly confident or top-ranked teacher predictions.
These areas are shown to be near instance centers or error-prone positions.
Besides, to facilitate effective knowledge transfer from teacher to student, we develop Teacher Guided Initialization (TGI) which remaps pre-trained teacher parameters to initialize a student model. This is shown to be effective in inheriting teacher models' feature extraction abilities while collaborating well with logit and label KD techniques.

Finally, our empirical studies on efficient model design and knowledge distillation methods yield superior performance in delivering efficient and effective pillar- and voxel-based 3D detectors. This is extensively verified on the largest annotated 3D object detection dataset -- Waymo~\cite{sun2020scalability}.
As shown in Figure~\ref{fig:performance}, our best performing model, \ie CP-Voxel-S, even outperforms its teacher model (\ie CP-Voxel) while has $2.4\times$ fewer flops. 
Moreover, our most efficient pillar-based model (\ie CP-Pillar-v0.64) can run 51 FPS with $58.89\%$ mAPH {while previous fastest voxel/pillar-based detector -- PointPillar runs 23 FPS with $57.03\%$ mAPH on an NVIDIA A100 GPU (see Sec.~\ref{sec:more_analysis}).} Our method is also shown to be effective in knowledge transfer from heavy two-stage object detectors to lightweight single-stage detectors. 
In addition, our method can generalize well to other settings such as KITTI dataset with SECOND as well as advance compression methods in Sec.~\ref{sec:generality}, other detectors in Sec.~\ref{sec:supp_discussion}, and even 3D semantic segmentation in Sec.~\ref{sec:supp_generality}.

\vspace{-0.3cm}
\section{Related Work}
\vspace{-0.3cm}
\textbf{3D LiDAR-based Object Detection} targets to localize and classify 3D objects from point clouds. Point-based methods ~\cite{shi2019pointrcnn,chen2019fast,yang2019std,zhang2022not} took raw point clouds and leveraged PointNet++~\cite{qi2017pointnet++} to extract sparse point features and generate point-wise 3D proposals. 
Pillar-based works~\cite{lang2019pointpillars,wang2020pillar} finished voxelization in bird eye's view and extracted pillar-wise features with PointNet++.
Voxel-based methods~\cite{zhou2018voxelnet,yan2018second,shi2020pv} voxelized point clouds and obtained voxel-wise features with 3D sparse convolutional networks, which is the most popular data treatment. 
Besides, range-based works~\cite{bewley2020range,sun2021rsn} were proposed for long-range and fast detection. 
Recently, designing efficient 3D detectors has drawn some attentions~\cite{chen2019fast,zhang2022not} with raw point data treatment. 
In this work, we focus on exploring model-agnostic knowledge distillation methods to boost the efficiency of 3D detectors.


\textbf{Knowledge Distillation} aims to transfer knowledge from a large teacher model to a lightweight student network, which is a thriving area in efficient deep learning. Hinton \etal proposed the seminal concept of knowledge distillation (KD)~\cite{hinton2015distilling}, which distilled knowledge between teacher and student on the output level (\ie prediction logits). Another line of research proposed to help student's optimization with hints stored in informative intermediate features from teacher~\cite{romero2014fitnets, huang2017like,komodakis2017paying,heo2019knowledge,jin2019knowledge,chen2021cross}. In addition, some works attempted distillation techniques in 2D object detection~\cite{li2017mimicking,wang2019distilling, dai2021general,qi2021multi,yang2021focal,nguyen2022improving} by emphasizing instance-wise distillation and feature knowledge.
Mimic~\cite{li2017mimicking}, FG~\cite{wang2019distilling} and GID~\cite{dai2021general} sampled local region features with box proposals or custom indicators for foreground-aware feature imitation. Label KD~\cite{nguyen2022improving} utilized teacher's information for label assignment of student.
Recently, knowledge distillation has also been leveraged to transfer knowledge in multi-modality setup~\cite{guo2021liga,liu20213d} or multi-frame to single-frame setup~\cite{wang2020multi} in 3D detection area. However, to the best of our knowledge, we are the first to explore knowledge distillation in the most popular setup: single-frame 3D LiDAR-based object detection. 
In this work, we propose an enhanced 3D detection KD pipeline with our designed efficient 3D detectors on the popular voxel/pillar data representation. Our lightweight detectors CP-Voxel-S and CP-Pillar-v0.4 slightly outperform their state-of-the-art teacher detectors separately while requiring much less computation overhead.
\vspace{-0.3cm}
\section{Designing Efficient Student Networks}
\label{sec:student_design}
\vspace{-0.3cm}
As there are no readily available lightweight backbone architectures for constructing student networks, we carry out an empirical study on how to obtain an efficient model with satisfactory efficiency and accuracy trade offs to facilitate further knowledge distillation. In this section, we will first describe our experimental setups and model evaluation metrics. Then, we study different strategies to obtain an efficient model and conduct an in-depth analysis on how to achieve good trade offs for pillar- and voxel-based architectures.

\vspace{-0.3cm}
\subsection{Basic Setups and Evaluation Metrics}
\vspace{-0.2cm}

\textbf{Basic setups.} For detector architectures, we focus on two variants of the state-of-the-art model CenterPoint~\cite{yin2021center}: CenterPoint-Pillar (CP-Pillar) and CenterPoint-Voxel (CP-Voxel), covering the most popular pillar- and voxel-based 3D detectors~\cite{yan2018second,zhou2018voxelnet,lang2019pointpillars,wang2020pillar,shi2020pv,shi2021pv,deng2020voxel}.
For dataset, we perform all experiments on the largest annotated 3D LiDAR perception dataset Waymo Open Dataset (WOD) \cite{sun2020scalability} with $20\%$ training samples for fast verification. 
For model training, we follow the training scheme of popular 3D detection codebase OpenPCDet~\cite{openpcdet2020} to ensure fair comparisons and standardization.
Note that there is no any knowledge distillation method engaged in this stage.


\textbf{Evaluation metrics.} 
%
Following ~\cite{radosavovic2020designing,li2021benchmarking}, we employ number of parameters, flops, activations, latency (\ie test time) and peak GPU training memory (batch size 1) as quantitative indicators to evaluate model efficiency from parameter, computation and memory throughout aspects. Note that activations, flops and latency are averaged over 99 frames with a GTX-1060 GPU. We present latency more for reference since it largely depends on hardware devices as well as operation-level optimizations (see Sec.~\ref{sec:more_analysis}). We use LEVEL 2 mAPH as the performance evaluation metric following WOD~\cite{sun2020scalability}.


Since our major target here is to design student networks with favorable trade offs between performance and efficiency, we propose a quantitative indicator, namely Cost Performance Ratio (CPR), to directly measure a model in this respective.
To construct CPR, we use activations (acts) as the  metric to evaluate efficiency since it strongly correlates with the runtime on hardware accelerators such as GPUs as shown in ~\cite{radosavovic2020designing}.
Our CPR finally combines the activation ratio and mAPH as follows:
\vspace{-0.3cm}
\begin{spacing}{0.5}
\begin{equation}
\begin{small}
    \text{CPR} = 0.5 \times (1 -  \frac{\text{acts}_s}{\text{acts}_t}) +  0.5 \times (\frac{\text{mAPH}_s}{\text{mAPH}_t})^3,
\end{small}
\end{equation}
\end{spacing}
where the subscripts $s$ and $t$ represent student and teacher model respectively. CPR is normalized to $[0, 1]$ by weighting the relative activation decrease and performance drop ratio of a student network compared to the teacher. Notice that the third power is used for the performance degradation term to penalize acceleration methods that result in drastic performance degradation more severely. We argue that it is necessary to ensure a relative good performance of efficient detectors as models with poor accuracy may suffer from architectural level problems which can cause difficulties in developing knowledge distillation techniques and prohibiting obtaining accurate and efficient detectors. 

\begin{figure}[t]
\vspace{-0.3cm}
\begin{center}
\includegraphics[width=1\linewidth]{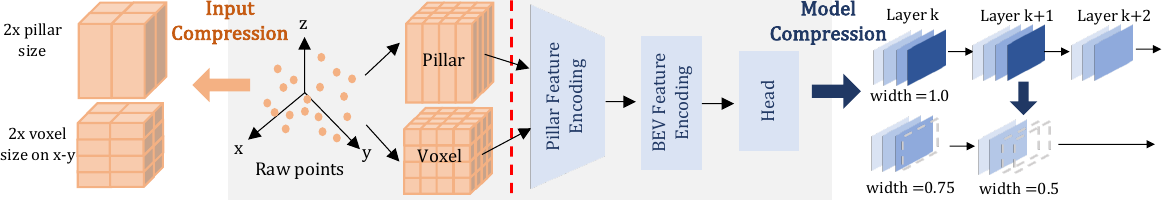}
\end{center}
\vspace{-0.5cm}
\caption{Architecture of detector and illustration of model (right) and input (left) compression.}
\vspace{-0.6cm}
\label{fig:compress}
\end{figure}

\vspace{-0.2cm}
\subsection{Acceleration Strategy}
\vspace{-0.2cm}
Here, to obtain efficient models, we investigate model and input resolution compression techniques on pillar- and voxel-based 3D detectors as shown in Figure~\ref{fig:compress} and study their impacts on model accuracy.

\textbf{Model Compression.}
Here, we compress a given teacher backbone by trimming it along depth (\ie number of layers) or width (number of channels) as shown in the right of Figure~\ref{fig:compress}.  
To relieve the burden of exhaustive layer-wise studies, we leverage the functional similarity among different layers to group them into three major modules: Pillar Feature Encoding (PFE) module, Bird eye's view Feature Encoding (BFE) module and detection head (see Figure~\ref{fig:compress}: right), and conduct analysis on module-level. Specifically, PFE corresponds to network components before projecting features to bird eye's view (BEV) grid, including sparse 3D convolutional backbone~\cite{graham2017submanifold} to extract voxel-wise features in voxel-based architectures or PointNet++~\cite{qi2017pointnet++} to encode pillar-wise features from points in pillar-based detectors.
After PFE, the features are aggregated to pillar features and mapped to 2D BEV grid. Then, the BFE consisting of 2D convolutional layers is adopted to extract final detection feature map on the BEV grid.
Finally, the detection head takes outputs from BFE to produce final prediction results.
%
For width pruning, we consider the width of teacher detectors as $1.0$, and slim each module of the student by decreasing its number of channels with a given width. Depth trimming also follows this paradigm but still maintains the minimal structure of each module (\eg downsample and upsample layers) to ensure basic detection ability. Model trimming results are shown in Table~\ref{tab:model_prune}.

\begin{table}[htbp]
    \centering
    \vspace{-0.3cm}
    \caption{Model compression results. Teacher models are marked in gray. See text for details.}
    \vspace{-0.3cm}
    \begin{small}
    \setlength\tabcolsep{3pt}
    \scalebox{0.97}{
        \begin{tabular}{cc|c|c|c|c|c|c|c|c|c|c|c|c}
            \bottomrule[1pt]
            \multicolumn{7}{c|}{Architecture} & \multicolumn{5}{c|}{Efficiency} & \multirow{3}{*}{\makecell[c]{LEVEL 2 \\ mAPH}} & \multirow{3}{*}{CPR} \\
            \cline{1-12}
            \multicolumn{2}{c|}{\multirow{2}{*}{Detector}} & \multicolumn{3}{c|}{Width} & \multicolumn{2}{c|}{Depth} &  Params  & Flops & Acts & Latency & Mem. & & \\
            \cline{3-7}
            & & PFE & BFE & Head & PFE & BFE & (M) & (G) & (M) & (ms) & (G) & &  \\
            \hline
            \cellcolor{white!25} \multirow{6}{*}{CP-Pillar} & \cellcolor{white!25} & \cellcolor{Gray!16} 1.00 & \cellcolor{Gray!16} 1.00 & \cellcolor{Gray!16} 1.00 & \cellcolor{Gray!16} 1.00 & \cellcolor{Gray!16} 1.00 & \cellcolor{Gray!16} 5.2 & \cellcolor{Gray!16} 333.9 & \cellcolor{Gray!16} 303.0 & \cellcolor{Gray!16} 157.9 & \cellcolor{Gray!16} 5.2 & \cellcolor{Gray!16} 59.09 & \cellcolor{Gray!16} - \\
            \cline{3-14}
            & (a) & \cellcolor{blue!10} 1.00 & \cellcolor{blue!10} 0.50 & \cellcolor{blue!10} 0.50 & 1.00 & 1.00 & 1.3 & 87.6 & 161.8 & 78.5 & 3.2 & 54.50 & 0.63 \\
            \cline{3-14}
            & (b) & \cellcolor{blue!10} 0.50 & \cellcolor{blue!10} 0.50 & \cellcolor{blue!10} 0.50 & 1.00 & 1.00 & 1.3 & 85.0 & 152.7 & 74.0 & 2.9 & 52.33 & 0.60 \\
            \cline{3-14}
            & (c) & 1.00 & 1.00 & 1.00 & \cellcolor{blue!10} 1.00 & \cellcolor{blue!10} 0.50 & 2.2 & 258.5 & 234.1 & 118.5 & 4.3 & 55.24 & 0.52 \\
            \cline{3-14}
            & (d) & 1.00 & 1.00 & 1.00 & \cellcolor{blue!10} 1.00 & \cellcolor{blue!10} 0.33  & 1.4 & 234.6 & 210.0 & 107.8 & 4.0 & 47.97 & 0.42 \\
            \toprule[1pt]
            \bottomrule[1pt]
            \cellcolor{white!25} \multirow{8}{*}{CP-Voxel} & \cellcolor{white!25} & \cellcolor{Gray!16} 1.00 & \cellcolor{Gray!16} 1.00 &  \cellcolor{Gray!16} 1.00 & \cellcolor{Gray!16} 1.00 & \cellcolor{Gray!16} 1.00 & \cellcolor{Gray!16} 7.8 & \cellcolor{Gray!16} 114.7 & \cellcolor{Gray!16} 101.9 & \cellcolor{Gray!16} 125.7 & \cellcolor{Gray!16} 2.8 & \cellcolor{Gray!16} 64.29 & \cellcolor{Gray!16} - \\
            \cline{3-14}
            & (a) & \cellcolor{blue!10} 1.00 & \cellcolor{blue!10} 0.50 & \cellcolor{blue!10} 0.50 & 1.00 & 1.00 & 4.0 & 47.8 & 65.7 & 98.0 & 2.1 & 62.23 & 0.63 \\
            \cline{3-14}
            & (b) & \cellcolor{blue!10} 0.75 & \cellcolor{blue!10} 0.50 & \cellcolor{blue!10} 0.50 & 1.00 & 1.00 & 2.8 & 36.9 & 58.4 & 88.2 & 1.9 & 61.16 & 0.64 \\
            \cline{3-14}
            & (c) & \cellcolor{blue!10} 0.50 & \cellcolor{blue!10} 0.50 & \cellcolor{blue!10} 0.50 & 1.00 & 1.00 & 1.9 & 28.8 & 51.2 & 75.1 & 1.7 & 59.47 & 0.64 \\
            \cline{3-14}
            & (d) & \cellcolor{blue!10} 0.50 & \cellcolor{blue!10} 0.25 & \cellcolor{blue!10} 0.25 & 1.00 & 1.00 & 1.0 & 12.0 & 33.1 & 70.4 & 1.3 & 56.26 & 0.67 \\
            \cline{3-14}
            & (e) & \cellcolor{blue!10} 0.25 & \cellcolor{blue!10} 0.25 & \cellcolor{blue!10} 0.25 & 1.00 & 1.00 & 0.5 & 7.3 & 25.8 & 66.0 & 1.1 & 49.84 & 0.61 \\
            \cline{3-14}
            & (f) & 1.00 & 1.00 & 1.00 & \cellcolor{blue!10} 0.50 & \cellcolor{blue!10} 0.50 & 3.0 & 63.9 & 65.2 & 73.0 & 1.9 & 60.95 & 0.61 \\
            \cline{3-14}
            & (g) & 1.00 & 1.00 & 1.00 & \cellcolor{blue!10} 0.33 & \cellcolor{blue!10} 0.33 & 1.8 & 47.9 & 52.2 & 59.0 & 1.6 & 55.78 & 0.57 \\
            \toprule[0.8pt]
        \end{tabular}
    }
    \end{small}
    \label{tab:model_prune}
    \vspace{-0.3cm}
\end{table}

\textbf{Input Compression.}
Besides model complexities, input resolution also has impacts on model efficiency~\cite{tan2019efficientnet,qi2021multi}.   
For instance, by halving the input resolution, the computation overhead for 2D convolution layers in the BFE module and detection head will be reduced to $\frac{1}{4}$ (see Figure~\ref{fig:compress}: left).
Besides, there are large background areas in the sparse and large-scale 3D scenes, which naturally has redundancies and offers the possibility of processing the data on a coarser resolution for input compression.
Specifically, input compression is realized by increasing the voxel/pillar size on the x-y plane when constructing voxel/pillar. As shown in Table~\ref{tab:input_compress}, we gradually increase students' voxel size with $25\%$ of teachers' voxel size, and record their efficiency and accuracy metrics.


\begin{table*}[htbp]
    \centering
    \vspace{-0.6cm}
    \caption{Input compression results. Teacher models are marked in gray. See text for details.}
    \begin{small}
    \setlength\tabcolsep{3.5pt}
    \scalebox{0.94}{
        \begin{tabular}{c|c|c|c|c|c|c|c|c}
            \bottomrule[1pt]
            \multicolumn{2}{c|}{Architecture} & \multicolumn{5}{c|}{Efficiency} & LEVEL 2 & \multirow{2}{*}{CPR} \\
            \cline{1-7}
            Detector & Voxel Size (m) & Params (M) & Flops (G) & Acts (M) & Latency (ms) & Mem. (G) & mAPH \\
            \hline
            \cellcolor{white!25} \multirow{5}{*}{CP-Pillar} & \cellcolor{Gray!16} 0.32 & \cellcolor{Gray!16} 5.2 & \cellcolor{Gray!16} 333.9 & \cellcolor{Gray!16} 303.0 & \cellcolor{Gray!16} 157.9 & \cellcolor{Gray!16} 5.2 & \cellcolor{Gray!16} 59.09 & \cellcolor{Gray!16} - \\
            \cline{2-9}
            & 0.40  & 5.2 & 212.9 & 197.7 & 103.4 & 3.8 & 57.55 & 0.64 \\
            \cline{2-9}
            & 0.48 & 5.2 & 149.4 & 142.3 & 81.9 & 3.0 & 56.27 & 0.70  \\
            \cline{2-9}
            & 0.56  & 5.2 & 109.9 & 109.0 & 66.3 & 2.6 & 54.45 & 0.71 \\
            \cline{2-9}
            & 0.64  & 5.2 & 85.1 & 88.0 & 54.5 & 2.1 & 52.81 & 0.71 \\
            \toprule[1pt]
            \bottomrule[1pt]
            \cellcolor{white!25} \multirow{5}{*}{CP-Voxel} & \cellcolor{Gray!16} 0.100 & \cellcolor{Gray!16} 7.8 & \cellcolor{Gray!16} 114.8 & \cellcolor{Gray!16} 101.9 & \cellcolor{Gray!16} 125.7 & \cellcolor{Gray!16} 2.8 & \cellcolor{Gray!16} 64.29 & \cellcolor{Gray!16} - \\
            \cline{2-9}
            & 0.125 & 7.8 & 77.5 & 70.1 & 99.9 & 2.2 & 61.55 & 0.59 \\
            \cline{2-9}
            & 0.150 & 7.8 & 53.9 & 50.0 & 84.3 & 1.8 & 58.14 & 0.62 \\
            \cline{2-9}
            & 0.175 & 7.8 & 44.3 & 41.2 & 74.1 & 1.5 & 55.99 & 0.63 \\
            \cline{2-9}
            & 0.200 & 7.8 & 32.9  & 31.4 & 67.5 & 1.3 & 52.80 & 0.62 \\
            \toprule[0.8pt]
        \end{tabular}
    }
    \end{small}
    \label{tab:input_compress}
    \vspace{-0.6cm}
\end{table*}

\vspace{-0.2cm}
\subsection{Conclusion and Analysis} 
\label{sec:design_analy}
\vspace{-0.2cm}

By analyzing the experimental results shown in Table~\ref{tab:model_prune} and Table~\ref{tab:input_compress}, we draw the following conclusions on efficient model design for pillar- and voxel-based detectors.

\textbf{Width \vs depth compression: width-level pruning is preferred.}
As shown in Table~\ref{tab:model_prune}, trimming networks on width generally achieves higher CPR than on depth for both CP-Pillar and CP-Voxel. 
For instance, with stronger performance, CP-Voxel (d) needs $1.5\times$ fewer acts and $4\times$ fewer flops compared to CP-Voxel (g). As backbones of 3D detectors are much shallower than their 2D counterparts (\eg only 19 convolution layers in CP-Pillar and 35 convolution layers in CP-Voxel), this renders depth compression more challenging and less scalable than width compression in 3D detection.

\textbf{Module-wise pruning selection: PFE module has the least redundancy to be reduced.}
Since PFE, BFE and detection head perform different functions, they might have their own redundancies in 3D detection. 
Comparing  CP-Voxel (c), (d) and (e) as well as CP-Pillar (a) and (b) in Table~\ref{tab:model_prune}, we find that less pruning on PFE achieves significantly higher CPR, demonstrating that network parameters in the PFE module is crucial for high performance and it is necessary to maintain network complexities in PFE modules.


\textbf{Favorable compression strategies for different detection architectures.}
Comparing results of different compression strategies in Table~\ref{tab:model_prune} and~\ref{tab:input_compress}, we find that pillar-based architecture (\ie CP-Pillar) is more suitable for input compression while voxel-based architecture (\ie CP-Voxel) prefers width-level compression. This difference mainly lies in different spatial redundancy of BEV features on these two detectors. 
Since the PFE module of voxel-based detector downsamples input resolution to ${1}/{8}$, the $0.1m$ input voxel size will be magnified to $0.8m$ on BEV features. Hence, less spatial redundancy could be further compressed for CP-Voxel. 
However, CP-Pillar often keeps the same resolution between input and BEV features on Waymo~\cite{openpcdet2020,yin2021center}, and thus has larger resolution redundancy on BEV grid, which facilitates designing student network with coarser input resolution.

\textbf{Summarized student networks.}
Driven by the above analysis, we finally derive the following compressed student models with good trade offs between efficiency and performance. For CP-Pillar, as it is friendly to input compression, we adopt the input-compressed models with voxel size 0.40, 0.48 and 0.64 in Table~\ref{tab:input_compress}, named \textbf{CP-Pillar-v0.4}, \textbf{CP-Pillar-v0.48} and \textbf{CP-Pillar-v0.64}, respectively.
As for CP-Voxel, we choose its width compressed models (a) (b) and (d) in Table~\ref{tab:model_prune}, named \textbf{CP-Voxel-S}, \textbf{CP-Voxel-XS} and \textbf{CP-Voxel-XXS}, respectively.


\vspace{-0.2cm}
\section{Benchmark Knowledge Distillation for 3D Object Detection}
\vspace{-0.2cm}
With student networks derived in Section~\ref{sec:student_design}, we are now ready to conduct our empirical study on knowledge distillation for 3D object detection. 
Here, we benchmark seven popular 2D KD methods including logit KD (\ie KD~\cite{hinton2015distilling} and GID-L~\cite{dai2021general}), feature KD (\ie FitNet~\cite{romero2014fitnets}, Mimic~\cite{li2017mimicking}, FG~\cite{wang2019distilling} and GID-F~\cite{dai2021general}) and label KD~\cite{nguyen2022improving} on six teacher-student pairs with comprehensive analysis. Notice that GID and label KD are state-of-the-art 2D KD detection methods, and GID is divided into GID-L and GID-F to investigate logit KD and feature KD separately. Implementation details and the value of hyper-parameters are described in the Sec.~\ref{sec:implmentation_details}.

\begin{figure}[h]
    \centering
    \vspace{-0.4cm}
    \includegraphics[width=1\linewidth]{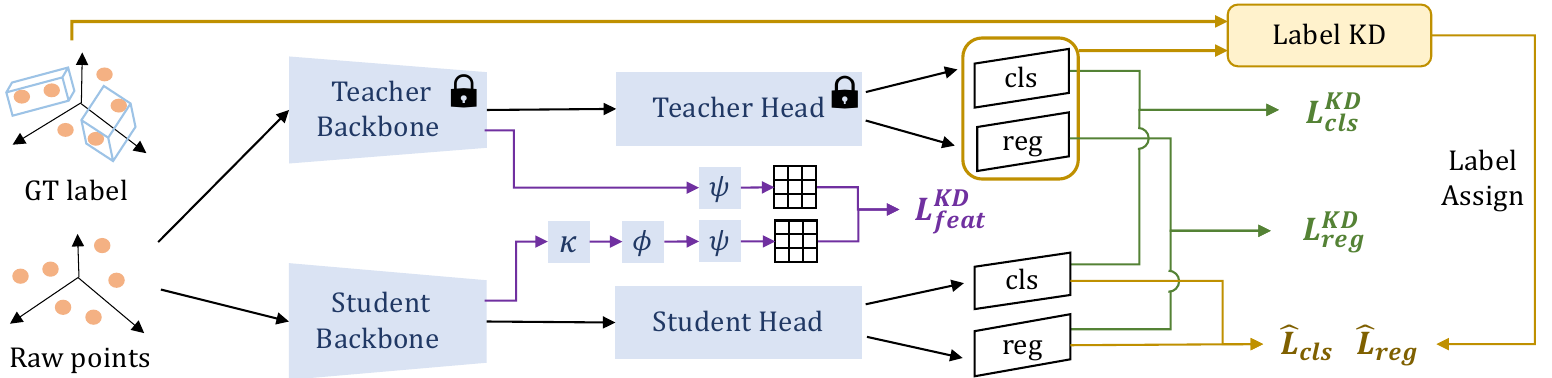}
    \vspace{-0.6cm}
    \caption{Overall KD paradigm for 3D detection. Teacher weights are frozen during the whole distillation procedure. Logit, feature and label KD are colored by green, purple and yellow, respectively.}
    \label{fig:paradigm}
    \vspace{-0.4cm}
\end{figure}

\subsection{Paradigm}
\vspace{-0.2cm}
We first evaluate existing 2D distillation methods for 3D detection. As shown in Figure~\ref{fig:paradigm}, our overall KD paradigm for 3D detection contains three parts: logit, feature and label KD to leverage teacher guidance in response, intermediate feature and label assignment levels, respectively. 

\textbf{Logit KD} is the most classical distillation approach introduced by \cite{hinton2015distilling}. It takes teacher model's final response as guidance for training a student network and is closely related to the specific task. In 3D object detection, we calculate the logit KD loss between teacher and student outputs as follows:
\vspace{-0.2cm}
\begin{spacing}{0.5}
\begin{equation}
\label{eq:logit_kd}
\begin{aligned}
\begin{small}
    \mathcal{L}^{\text{KD}}_{\text{cls}} = \mathbb{E}[m_{\text{cls}} \| \kappa(p^s_{\text{cls}}) - p^t_{\text{cls}} \|_2], \ \ \ \ \ \ \ \ \ \ \ \  \mathcal{L}^{\text{KD}}_{\text{reg}} = \mathcal{L}_{\text{reg}}(p^s_{\text{reg}},\  p^t_{\text{reg}}),
\end{small}
\end{aligned}
\end{equation}
\end{spacing}
where superscripts $s$ and $t$ indicate student and teacher, $p_{\text{cls}}$ and $p_{\text{reg}}$ represent the classification response after the sigmoid and bounding box regression prediction of detector separately, $\kappa$ is the bilinear interpolation to match student output resolutions towards teacher, $\mathcal{L}_{\text{reg}}$ is the regression loss function of 3D detector and $m_{\text{cls}}$ is a mask ranged in $[0, 1]$ to indicate important regions in $p_{\text{cls}}$ (see green parts in Figure~\ref{fig:paradigm}).
Compared to vanilla KD~\cite{hinton2015distilling} which mimics all teacher outputs, GID-L only focuses on some local regions covered by selected box proposals and results in different $m_{\text{cls}}$.

\textbf{Feature KD} is the major stream of work in KD for 2D object detection. It enforces student models to mimic teacher models' intermediate feature maps. Specifically, we construct feature mimicking on the last layer of BFE between student and teacher network as follows:
\vspace{-0.2cm}
\begin{spacing}{0.6}
\begin{equation}
\begin{small}
\mathcal{L}^{\text{KD}}_{\text{feat}} = \mathbb{E}[ m_{\text{feat}} \| \psi( \phi(\kappa(f^s)), y) - \psi(f^t, y) \|_2],
\end{small}
\end{equation}
\end{spacing}
where $y$ is the ground truth, $\psi$ indicates the RoI Align~\cite{he2017mask}, $\phi$ is a $1\times1$ convolution with batch normalization~\cite{ioffe2015batch} and ReLU~\cite{nair2010rectified} block to align channel-wise discrepancy between teacher feature $f^t$ and student feature $f^s$, $m_{\text{feat}}$ is the mask to indicate critical regions ranged in $[0, 1]$ (see purple parts in Figure~\ref{fig:paradigm}). 
Considering the imbalance between foreground and background regions in the detection, a prevailing solution is to emphasize near object regions to perform distillation~\cite{li2017mimicking,wang2019distilling,dai2021general}.
Different investigated feature KD methods mainly vary in the selection of such critical regions $m_{\text{feat}}$.

\textbf{Label KD} is a newly proposed distillation strategy, which leverages teacher predictions in the label assignment stage of student ~\cite{nguyen2022improving}. Motivated by the simple and general label KD, we also employ it as a KD baseline method. Specifically, given a point cloud scene $x$ and its corresponding ground truth (GT) set $y$, label KD first obtains the prediction $y^t$ and confidence score $o^t$ from the  pretrained teacher detector. After filtering $o^t$ with a given threshold $\tau$, it then obtains a high-quality teacher prediction set $\hat{y}^t$. It generates a teacher assisted GT set $\hat{y}^{\text{KD}} = \{y, \hat{y}^t\}$ by combining GTs as well as confident teacher predictions and carries on label assignment for student with $\hat{y}^{\text{KD}}$. The final classification and regression losses with label KD on student detectors are $\mathcal{\hat{L}}_{\text{cls}}$ and $\mathcal{\hat{L}}_{\text{reg}}$. 

\textbf{Training Objective} is the combination of three stream KD techniques as follows:
\vspace{-0.3cm}
\begin{spacing}{0.5}
\begin{equation}\label{eq:objective}
\begin{small}
    \mathcal{L} = \mathcal{\hat{L}}_{\text{cls}} + \lambda\mathcal{\hat{L}}_{\text{reg}} + \alpha_1 \mathcal{L}^\text{KD}_{\text{cls}} + \alpha_2 \mathcal{L}^\text{KD}_{\text{reg}} + \alpha_3 \mathcal{L}^\text{KD}_{\text{feat}},
\end{small}
\end{equation}
\end{spacing}
where $\lambda$, $\alpha_1$, $\alpha_2$ and $\alpha_3$ are trade-off parameters between different objectives. 
Note that the existence and implementation of each operation (\eg $\psi$, $\kappa$, $m_{\text{feat}}$, $m_{\text{cls}}$, etc) varies among different KD methods.

\begin{table}[htbp]
    \centering
    \vspace{-0.2cm}
    \caption{Knowledge distillation benchmark for 3D detection on Waymo. Performance are measured in LEVEL 2 mAPH. Best and second-best methods are noted by \textbf{bold} and \underline{underline}, respectively. ``Ours'' indicates our proposed improved knowledge distillation method introduced in Sec.~\ref{sec:sparsekd}}
    \vspace{-0.2cm}
    \begin{small}
    \setlength\tabcolsep{3.7pt}
    \scalebox{1.0}{
        \begin{tabular}{l|c||c|c|c|c|c|c|c||c||c|c}
            \bottomrule[1pt]
            \multirow{2}{*}{Detector} & \multirow{2}{*}{\makecell[c]{No \\ Distill}} & \multicolumn{2}{c|}{Logit KD} & \multicolumn{4}{c|}{Feature KD} & \multirow{2}{*}{\makecell[c]{Label \\ KD}} & \multirow{2}{*}{Ours} & Flops & Acts \\
            \cline{3-8}
             & & KD & GID-L & FitNet & Mimic & FG & GID-F & & & (G) & (M)  \\
            \hline
            \rowcolor{Gray!16} CP-Pillar & 59.09 & - & - & - & - & - & - & - & - & 333.9 & 303.0 \\
            CP-Pillar-v0.4 & 57.55 & 57.51 & 57.54 & 57.89 & \underline{58.57} & 58.44 & 58.26 & 58.10 & \textbf{59.24} & 212.9 & 197.7 \\
            CP-Pillar-v0.48 & 56.27 & 55.76 & 56.29 & 55.82 & 57.26  & 57.26 & 57.23 & \underline{57.54} & \textbf{58.53} & 149.4 & 142.3 \\
            CP-Pillar-v0.64 &  52.81 & 53.13 & 50.78 & 51.79 & \underline{53.83} & 53.37 & 53.18 & 53.78 & \textbf{55.82} & 85.1 & 88.0 \\
            \toprule[1pt]
            \bottomrule[1pt]
            \rowcolor{Gray!16} CP-Voxel & 64.29 & - & - & - & - & - & - & - & - & 114.7 & 101.9 \\
            CP-Voxel-S & 62.23 & 62.81 & 62.89 & 60.51 & \underline{63.35} & 63.33 & 62.75 & 63.31 & \textbf{64.25} & 47.8 & 65.7 \\
            CP-Voxel-XS & 61.16 & 61.30 & 62.25 & 58.94 & 62.23 & \underline{62.48} & 62.42 & 61.81 & \textbf{63.53} & 36.9 & 58.4 \\
            CP-Voxel-XXS & 56.26 & 56.11 & 57.19 & 52.24 & 57.00 & \underline{57.92} & 57.16 & 57.02 & \textbf{59.28} & 12.0 & 33.1 \\
            \toprule[0.8pt]
        \end{tabular}
    }
    \end{small}
    \label{tab:kd_benchmark}
    \vspace{-0.3cm}
\end{table}
\vspace{-0.2cm}
\subsection{Results and Analysis}
\label{sec:analysis_of_kd_benchmark}
\vspace{-0.2cm}
\textbf{Benchmark Analysis.} As shown in Table~\ref{tab:kd_benchmark}, compared to the no distillation baseline, all three streams of KD methods obtain performance improvements on six teacher-student pairs. 
Among seven KD baseline strategies, feature-based KD methods (\ie Mimic and FG) achieve prominent performance, which demonstrates the strong potential of learning from teacher's hints on feature extraction. 
Furthermore, we find that instance-aware local region imitation is important in distillation for 3D detection, as enormous background regions overwhelm the supervision of sparse instances. For instance, with instance-aware imitation, Mimic, FG and GID-F consistently outperform FitNet which fully imitates all spatial positions of teacher feature maps. 
Similar conclusions can also be drawn in logit KD by comparing the results of instance-aware GID-L and vanilla KD. 

\begin{wraptable}{r}{4.3cm}
    \vspace{-0.7cm}
    \centering
    \caption{Synergy investigation based on CP-Voxel-XXS.}
    \begin{small}
    \setlength\tabcolsep{3pt}
    \scalebox{0.9}{
        \begin{tabular}{ccc|c}
            \bottomrule[1pt]
            GID-L & Label KD & FG & mAPH \\
            \hline
             & & & 56.26 \\
            \hline
             $\surd$ & & & 57.19 \\
             & $\surd$ & & 57.02 \\
             & & $\surd$ & 57.92 \\
            \hline
             $\surd$ & $\surd$ &  & 57.60 \\
             $\surd$ &  & $\surd$ & \textbf{58.01} \\
             & $\surd$ & $\surd$ & 57.06 \\
            $\surd$ & $\surd$ & $\surd$ & 57.62 \\
            \toprule[0.8pt]
        \end{tabular}
    }
    \end{small}
    \label{tab:synergy_analy}
    \vspace{-0.5cm}
\end{wraptable}

\textbf{Synergy Analysis.}
While benchmark results in Table~\ref{tab:kd_benchmark} mainly focus on the individual effectiveness of each KD manner, their synergy effect is also an important consideration, which has the potential to further improve student performance. As shown in Table~\ref{tab:synergy_analy}, although feature KD itself achieves the highest performance on CP-Voxel-XXS compared to logit KD and label KD techniques, it can hardly achieve improvements or even suffers from performance degradation when combined with other KD methods. On the contrary, logit KD and label KD can cooperate well with each other to further improve student's capability. This is potentially caused by logit KD and label KD implicitly enforcing regularization on the feature, which can be conflicted with the optimization direction of feature KD and results in a poor synergy effect.

\vspace{-0.3cm}
\section{Improved Knowledge Distillation for 3D Object Detection}
\vspace{-0.3cm}
\label{sec:sparsekd}
As shown in Table~\ref{tab:synergy_analy}, the basic knowledge distillation pipeline fails to achieve remarkable results by combining the best method of three KD streams (\ie  GID-L~\cite{dai2021general}, label KD~\cite{nguyen2022improving} and FG~\cite{wang2019distilling}). In the following, we propose an improved KD pipeline for 3D object detection, including pivotal position logit KD to alleviate the extreme imbalance between foreground and background regions through only imitating response on sparse pivotal positions, label KD, and teacher guided initialization scheme to further facilitate transferring teacher's feature extraction ability to student models.

\vspace{-0.2cm}
\subsection{Method}
\vspace{-0.2cm}
\label{sec:method}
\textbf{Pivotal Position Logit KD.}
Motivated by the imbalance of foreground and background regions, previous 2D methods attempt to only enforce output-level imitation on pixels near or covering instances~\cite{wang2019distilling,dai2021general}. However, we find that it is sub-optimal in 3D scenarios given more extreme imbalance between small informative instances and large redundant background areas. 
For example, based on CP-Pillar, even a vehicle with 10$m$ length and 4$m$ width  occupies only $32\times 13$ pixels in the final $468 \times 468$ response map. 
Such small instances and large perception ranges in 3D detection requires more sophisticated imitation region selection than previous coarse instance-wise masking manners in 2D detection.
Hence, we propose Pivotal Position (PP) logit KD which leverages cues in teacher classification response or label assignment to determine the important areas for distillation.

%

Specifically, pivotal position selection can be formulated as finding suitable $m_{\text{cls}}$ in Eq.~\eqref{eq:logit_kd}. Here, we show three variants to obtain it. 
First, confidence of high-performance teacher prediction can serve as a valuable indicator to figure out pivotal positions for student (\ie confidence PP). By filtering teacher confidence $o^t$ with a threshold $\tau_{pp}$, we then set $i, j$ positions of $m_{\text{cls}}$ with $o^t_{i,j} \geq \tau_{pp}$ to one and otherwise zero. With similar spirit, we can also select top-ranked $K$ positions (\ie rank PP) in teacher classification response $P^t_{\text{cls}}$ as pivotal positions and convert them to a one-hot embedded $m_{\text{cls}}$. These confident or top-ranked positions are shown to be near object centers or error-prone regions.
Last, inspired by the Gaussian label assignment in CenterPoint~\cite{yin2021center}, we can define pivotal positions in a soft way with center-peak Gaussian distribution for each instance as $m_{\text{cls}}$ (\ie Gaussian PP). We empirically show that all three variants of PP logit KD achieve promising gains in 3D detection.

\begin{figure}[h]
    \centering
    \vspace{-0.2cm}
    \includegraphics[width=1\linewidth]{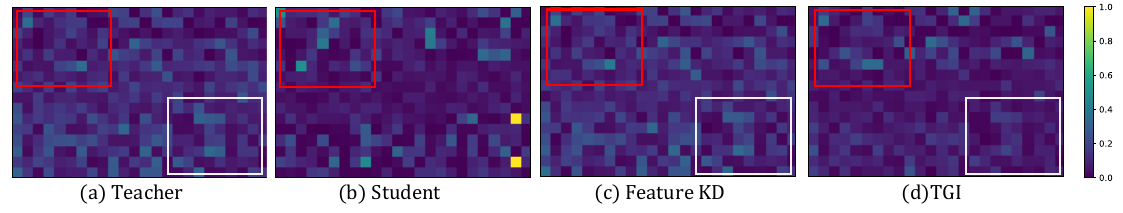}
    \vspace{-0.8cm}
    \caption{Visualization of channel-wise $L_1$ norm for distillation from CP-Pillar to CP-Pillar-v0.64.}
    \label{fig:ch_mag}
    \vspace{-0.2cm}
\end{figure}

\textbf{Teacher Guided Initialization.}
As mentioned in Section~\ref{sec:analysis_of_kd_benchmark}, feature KD itself outperforms other KD methods, but its deficient synergy results hinder 3D detection KD pipeline from achieving promising performance. This might be caused by the conflict between different optimization directions on intermediate features from five KD loss terms in Eq.~\eqref{eq:objective}.
Hence, we explore whether there is a substitute manner which can also leverage teacher's guidance on the feature extraction aspect.
%
Since the value of feature map is determined by model's weights, finding hints from teacher weights becomes an alternative to guide student's feature extraction.

Hence, we propose to directly use the trained weights of teacher to serve as the initialization of student network, named teacher guided initialization (\textbf{TGI}) to enhance student model's feature extraction abilities by inheriting it from a teacher model.
Although such initialization is intuitive for input compressed students (\eg CP-Pillar-v0.64), it cannot be directly applied to pruned students (\eg CP-Voxel-S). 
Therefore, we employ the parameter remapping strategy FNA~\cite{fang2020fast} to project teacher weights on student parameters. 
Take 2D convolution layer as an example, teacher and student parameters are represented with $W^t \in \mathbb{R}^{r \times q \times h \times w}$ and $W^s \in \mathbb{R}^{v \times u \times h \times w}$, where $v \leq r$ and $u \leq q$, respectively. 
The first $v$ and $u$ channels of teacher parameters are directly assigned to the slimmed student network in FNA. 
We visualize the channel-wise $L_1$ norm of backbone feature to compare TGI extracted features with other strategies.
As shown in Figure~\ref{fig:ch_mag}, our TGI extracts feature similar to teacher's (see red box region), but not naively mimics all teacher channels as feature KD does (see white box region). We argue that fully imitating all teacher features might penalize student optimization regarding the architecture or input resolution discrepancy between teacher-student pairs.
We empirically show that our simple TGI strategy achieves comparable or even better performance individually among four feature KD methods and shows prominent synergy effects on six teacher-student pairs (See Sec.~\ref{sec:extra_exper}).

\vspace{-0.2cm}
\subsection{Main Results and Comparison}
\vspace{-0.2cm}
To verify our improved KD pipeline, we conduct experiments on all six teacher-student pairs. As shown in Table~\ref{tab:kd_benchmark}, our improved KD pipeline consistently surpasses previous KD strategies on all settings with $0.7\% \sim 1.9\%$ improvement, thanks to our enhanced logit KD and more collaborative TGI.
Equipping our lightweight detectors with the improved KD pipeline, CP-Voxel-S obtains comparable performance to CP-Voxel in terms of mAPH with around $2.4\times$ fewer flops and $1.6\times$ fewer activations. Furthermore, with $3.1\times$ fewer flops, CP-Voxel-XS only suffers 0.76\% performance degradation. In addition, with around $1.6\times$ fewer flops and $1.5\times$ fewer activations, our distilled CP-Pillar-v0.4 even slightly outperforms CP-Pillar. CP-Pillar-v0.64 requires only $25\%$ flops and $29\%$ activations of teacher model, while achieving $55.82\%$ mAPH, only 3.27\% performance drop compared to CP-Pillar. 
These experimental results demonstrate that knowledge distillation is a promising technique to improve the performance of efficient 3D detectors.

\begin{wraptable}{r}{7cm}
    \centering
    \vspace{-1.2cm}
    \caption{Comparison with other detectors on full WOD. $\dag$ indicates results re-implemented by us.}
    \begin{small}
    \setlength\tabcolsep{1.2pt}
    \scalebox{0.85}{
        \begin{tabular}{l|cccc|c}
            \bottomrule[1pt]
            \multirow{2}{*}{Detector} & Params & Flops  & Acts & Latency & LEVEL 2 \\
            & (M) & (G) & (M) & (ms) & mAPH \\
            \hline
            PointPillar$\dag$~\cite{lang2019pointpillars} & 4.8 & 255.0 & 233.5 & 129.1 & 57.05 \\
            SECOND$\dag$~\cite{yan2018second} & 5.3 & 84.5 & 76.4 & 84.6 & 57.23 \\
            CP-Pillar$\dag$~\cite{yin2021center} & 5.2 & 333.9 & 303.0 & 157.9 & 61.56 \\
            CP-Voxel$\dag$~\cite{yin2021center} & 7.8 & 114.8 & 101.9 & 125.7 & 65.58 \\
            \hline
            PV-RCNN~\cite{shi2020pv} & 13.1 & 117.7 & 399.4 & 623.2 & 63.33 \\
            PV-RCNN++$\dag$~\cite{shi2021pv} & 16.1 & 123.5 & 179.7 & 435.9 & \textbf{69.46} \\
            \toprule[0.5pt]
            \bottomrule[0.5pt]
            CP-Voxel-S + Ours & 4.0 & 47.8 & 65.7 & 98.0 & \textbf{65.75} \\
            CP-Voxel-XS + Ours & 2.8 & 36.9 & 58.4 & 88.1 &  64.83 \\
            CP-Voxel-XXS + Ours & 1.0 & 12.0 & 33.1 & 70.4 &  60.93 \\
            \hline
            CP-Pillar-v0.4 + Ours &  5.2 & 212.9 & 197.7 & 103.4 & 61.60 \\
            CP-Pillar-v0.48 + Ours & 5.2 & 149.4 & 142.3 & 81.9 & 60.95 \\
            CP-Pillar-v0.64 + Ours & 5.2 &  85.1 & 88.0 & 54.5 & 58.89 \\
            \toprule[0.8pt]
        \end{tabular}
    }
    \end{small}
    \label{tab:other_detectors}
    \vspace{-0.7cm}
\end{wraptable}
\vspace{-0.2cm}
\subsection{Comparison with Other Detectors}
\vspace{-0.2cm}
To further demonstrate the efficiency and effectiveness of our designed detectors and KD pipeline, we also compare our distilled students with other detectors on the full WOD. As shown in Table~\ref{tab:other_detectors}, our CP-Voxel-S even  outperforms its teacher CP-Voxel with around $2.0\times$ fewer parameters, $2.4\times$ fewer flops and $1.6\times$ activations. With similar latency and much fewer parameters, flops as well as activations, CP-Voxel-XS outperforms SECOND by $7.6\%$. Our CP-Pillar-v0.64, is $2.4\times$ faster than PointPillar on a GTX-1060 while achieves $1.8\%$ higher performance.

\vspace{-0.1cm}
\subsection{Cross Stage Distillation}
\vspace{-0.2cm}
\begin{wraptable}{r}{7cm}
    \vspace{-0.9cm}
    \centering
    \caption{Cross stage 3D detector distillation.}
    \begin{small}
    \setlength\tabcolsep{1pt}
    \scalebox{0.85}{
        \begin{tabular}{c|c|c|c|c|c|c}
            \bottomrule[1pt]
            \multirow{2}{*}{Detector} & Params  & Flops & Acts & Latency & Mem. & LEVEL 2 \\
            & (M) & (G) & (M) & (ms) & (G) & mAPH \\
            \hline
            PV-RCNN++ & 16.1 & 123.5 & 179.7 & 435.9 & 4.2 & 67.80 \\
            CP-Voxel & 7.8 & 114.8 & 101.9 & 125.7 & 2.8 & 64.29 \\
            \hline
            CP-Voxel + Ours & 7.8 & 114.8 & 101.9 & 125.7 & 2.8 & \textbf{65.27} \\
            \toprule[0.8pt]
        \end{tabular}
    }
    \end{small}
    \label{tab:cross_stage}
    \vspace{-0.5cm}
\end{wraptable}
A prevailing strategy to improve state-of-the-art 3D object detectors is to adopt a object proposal refinement head for two stage detection~\cite{shi2020pv,deng2020voxel,shi2021pv}. However, despite performance improvements by $3.5\%$, PVRCNN++ requires around $2.2\times$ parameters, $1.8\times$ activations and $3.5\times$ latency compared to CP-Voxel (see Table~\ref{tab:cross_stage}). 
Such computation and parameter overheads hinder the real-world applications of state-of-the-art two-stage 3D detectors. Therefore, here we also investigate whether the knowledge of the two-stage detector can help the learning of single-stage detector. It is noteworthy that this is the first attempt at cross stage distillation in both 2D and 3D object detection.
As shown in Table~\ref{tab:cross_stage}, leveraging hints from pretrained PVRCNN++, our distilled CP-Voxel achieves around $1\%$ performance gains without any extra computation and parameter overheads during inference.

\vspace{-0.2cm}
\subsection{Generalization to More Scenarios}
\vspace{-0.2cm}
\label{sec:generality}
To demonstrate the generality of our compression and knowledge distillation manners, we provide experiments on other dataset, detector and compression manners. Besides, we construct experiment and discussion on extending our method to other task and more detectors in the Sec.~\ref{sec:supp_generality} and \ref{sec:supp_discussion}.

\textbf{Generality on Other Dataset and Detector.} 
As shown in Table~\ref{tab:kitti_compression} and Table~\ref{tab:kitti_kd}, both our compression conclusion in Sec.~\ref{sec:design_analy} and the improved KD method can generalize well to KITTI~\cite{geiger2012we} dataset with a new anchor-based detector SECOND. Especially, the SECOND (a) surpasses teacher performance by around 0.5\% with 3.5$\times$ fewer flops, showing the generality of our conclusion and methods.

\begin{table}[htbp]
    \centering
    \vspace{-0.2cm}
    \caption{Model and input compression results on KITTI. Teacher models are marked in gray.}
    \vspace{-0.3cm}
    \begin{small}
    \scalebox{0.99}{
        \begin{tabular}{cc|c|c|c|c|c|c|c|c|c}
            \bottomrule[1pt]
            \multicolumn{5}{c|}{Architecture} & \multicolumn{4}{c|}{Efficiency} & \multirow{3}{*}{\makecell[c]{Moderate \\ mAP@R40}} & \multirow{3}{*}{CPR} \\
            \cline{1-9}
            \multicolumn{2}{c|}{\multirow{2}{*}{Detector}} & \multicolumn{2}{c|}{Width} & Voxel Size &  Params  & Flops & Acts & Latency & & \\
            \cline{3-4}
            & & PFE & BFE & (m) & (M) & (G) & (M) & (ms) & &  \\
            \hline
            \cellcolor{white!25} \multirow{6}{*}{SECOND} & \cellcolor{white!25} & \cellcolor{Gray!16}1.00  & \cellcolor{Gray!16}1.00 & \cellcolor{Gray!16}0.05 & \cellcolor{Gray!16}5.3 & \cellcolor{Gray!16}80.5 & \cellcolor{Gray!16}69.3 & \cellcolor{Gray!16}77.4 & \cellcolor{Gray!16}67.24 & \cellcolor{Gray!16}- \\
            \cline{3-11}
            & (a) & \cellcolor{blue!10}0.75 & \cellcolor{blue!10}0.50 & 0.05 & 1.6 & 23.0 & 38.0 & 51.8 & 65.62 & 0.69 \\
            \cline{3-11}
            & (b) & \cellcolor{blue!10}0.50 & \cellcolor{blue!10}0.50 & 0.05 & 1.4 & 20.5 & 35.9 & 46.1 & 64.21 & 0.68 \\
            \cline{3-11}
            & (c) & \cellcolor{blue!10}0.50 & \cellcolor{blue!10}1.00 & 0.05 & 4.6 & 72.4 & 65.2 & 70.6 & 65.70 & 0.50 \\
            \cline{3-11}
            & (d) & 1.00 & 1.00 & \cellcolor{blue!10}0.10 & 5.3 & 21.2 & 19.4 & 34.2 & 54.32 & 0.62 \\
            \toprule[0.8pt]
        \end{tabular}
    }
    \end{small}
    \label{tab:kitti_compression}
    \vspace{-0.3cm}
\end{table}

\begin{table}[htbp]
    \centering
    \vspace{-0.2cm}
    \caption{Knowledge distillation results for 3D detection on KITTI. Performance are measured in moderate mAP over 40 recall positions. Best method is noted by \textbf{bold}.}
    \vspace{-0.2cm}
    \begin{small}
    \setlength\tabcolsep{4.2pt}
    \scalebox{1.0}{
        \begin{tabular}{l|c||c|c|c|c|c|c|c||c||c|c}
            \bottomrule[1pt]
            \multirow{2}{*}{Detector} & \multirow{2}{*}{\makecell[c]{No \\ Distill}} & \multicolumn{2}{c|}{Logit KD} & \multicolumn{4}{c|}{Feature KD} & \multirow{2}{*}{\makecell[c]{Label \\ KD}} & \multirow{2}{*}{Ours} & Flops & Acts \\
            \cline{3-8}
             & & KD & GID-L & FitNet & Mimic & FG & GID-F & & & (G) & (M)  \\
            \hline
            \rowcolor{Gray!16} SECOND & 67.24 & - & - & - & - & - & - & - & - & 80.5 & 69.3 \\
            SECOND (a) & 65.62 & 66.06 & 66.34 & 66.00 & 66.37 & 66.58 & 66.75 & 67.03 & \textbf{67.70} & 23.0 & 38.0 \\
            \toprule[0.8pt]
        \end{tabular}
    }
    \end{small}
    \label{tab:kitti_kd}
    \vspace{-0.3cm}
\end{table}

\textbf{Generality on Advance Compression Manner.} 
As shown in Table~\ref{tab:integrated_compress}, since the coarser-resolution detector has more architecture-level redundancy with less input information, CP-Pillar (f) achieves higher CPR by combining the model and input compression strategies.
Besides, after applying KD methods to CP-Pillar (f) as Table~\ref{tab:integrated_kd}, it is still more efficient though fewer improvements from KD strategies with less redundancy. 
This demonstrates that our pipeline can substantially obtain a more efficient detector by harvesting the progress of advanced compression and distillation manners. 

\begin{table}[htbp]
    \centering
    \vspace{-0.1cm}
    \caption{Integrating different compression manners on Waymo. Teacher model is marked in gray.}
    \vspace{-0.3cm}
    \begin{small}
    \setlength\tabcolsep{5pt}
    \scalebox{0.97}{
        \begin{tabular}{c|c|c|c|c|c|c|c|c|c|c}
            \bottomrule[1pt]
            \multicolumn{5}{c|}{Architecture} & \multicolumn{4}{c|}{Efficiency} & \multirow{3}{*}{\makecell[c]{LEVEL 2 \\ mAPH}} & \multirow{3}{*}{CPR} \\
            \cline{1-9}
            \multirow{2}{*}{Detector} & \multicolumn{3}{c|}{Width} & Voxel Size &  Params  & Flops & Acts & Latency & & \\
            \cline{2-4}
            & PFE & BFE & Head & (m) & (M) & (G) & (M) & (ms) & &  \\
            \hline
            \cellcolor{Gray!16}CP-Pillar & \cellcolor{Gray!16}1.00  & \cellcolor{Gray!16}1.00 & \cellcolor{Gray!16}1.00 & \cellcolor{Gray!16}0.32 & \cellcolor{Gray!16}5.2 & \cellcolor{Gray!16}333.9 & \cellcolor{Gray!16}303.0 & \cellcolor{Gray!16}157.9 & \cellcolor{Gray!16}59.09 & \cellcolor{Gray!16}- \\
            \cline{3-11}
             CP-Pillar-v0.4 & 1.00 & 1.00 & 1.00 & \cellcolor{blue!10}0.40 & 5.2 & 212.9 & 197.7 & 103.4 & 57.55 & 0.64 \\
            \cline{3-11}
            CP-Pillar (e) & \cellcolor{blue!10}1.00 & \cellcolor{blue!10}0.875 & \cellcolor{blue!10}0.875 & 0.32 & 4.0 & 260.1 & 267.7 & 134.7 & 58.53 & 0.54 \\
            \cline{3-11}
             CP-Pillar (f) & \cellcolor{blue!10}1.00 & \cellcolor{blue!10}0.875 & \cellcolor{blue!10}0.875 & \cellcolor{blue!10}0.40 & 4.0 & 163.9 & 175.5 & 92.1 & 57.36 & 0.67 \\
            \toprule[0.8pt]
        \end{tabular}
    }
    \end{small}
    \label{tab:integrated_compress}
    \vspace{-0.3cm}
\end{table}

\begin{table}[htbp]
    \centering
    \caption{Knowledge distillation results for more sophisticated compressed 3D detector on Waymo. Teacher model is marked by gray Best method is noted by \textbf{bold}. $^{\S}$ indicates the CPR is calculated according to the performance of best distilled student.}
    \vspace{-0.2cm}
    \begin{small}
    \setlength\tabcolsep{3.2pt}
    \scalebox{1.0}{
        \begin{tabular}{l|c||c|c|c|c|c|c|c||c||c|c|c}
            \bottomrule[1pt]
            \multirow{2}{*}{Detector} & \multirow{2}{*}{\makecell[c]{No \\ Distill}} & \multicolumn{2}{c|}{Logit KD} & \multicolumn{4}{c|}{Feature KD} & \multirow{2}{*}{\makecell[c]{Label \\ KD}} & \multirow{2}{*}{Ours} & Flops & Acts & \multirow{2}{*}{CPR$^{\S}$} \\
            \cline{3-8}
             & & KD & GID-L & FitNet & Mimic & FG & GID-F & & & (G) & (M)  \\
            \hline
            \rowcolor{Gray!16}CP-Pillar & 59.09 & - & - & - & - & - & - & - & - & 333.9 & 303.0 & - \\
            \hline
            CP-Pillar-v0.4 & 57.55 & 57.51 & 57.54 & 57.89 & 58.57 & 58.44 & 58.26 & 58.10 & \textbf{59.24} & 212.9 & 197.7 & 0.68 \\
            CP-Pillar (f) & 57.36 & 56.93 & 56.70 & 57.15 & 57.81 & 57.48 & 57.77 & 57.57 & \textbf{58.62} & 163.9 & 175.5 & 0.70 \\
            \toprule[0.8pt]
        \end{tabular}
    }
    \end{small}
    \label{tab:integrated_kd}
    \vspace{-0.2cm}
\end{table}

\vspace{-0.2cm}
\section{Ablation Studies}
\label{sec:ablation}
\vspace{-0.3cm}
In this section, we conduct extensive ablation experiments to in-depth investigate the effectiveness of each component in our improved KD pipeline.

\textbf{Component Ablation.}
Here, we investigate each component of our KD method and their synergy results. As shown in Table~\ref{tab:abl_component}, based on CP-Voxel-XXS, both PP logit KD and TGI can obtain around $1.4\%$ improvements separately. When incorporating with each other and label KD, they further obtain around $1.6\%$ gains, showing the prominent synergy impact of our components. On the contrary, feature KD even suffers $0.2 \sim 0.6\%$ performance drop when combined with other KD techniques.

\begin{table}[htbp]
    \centering
    \vspace{-0.1cm}
    \makebox[0pt][c]{\parbox{\textwidth}{%
    \begin{minipage}[t]{0.45\hsize}
        \centering
        \caption{Component ablation study based on CP-Voxel-XXS.}
        \vspace{-0.2cm}
        \begin{small}
        \setlength\tabcolsep{3pt}
        \scalebox{0.9}{
            \begin{tabular}{cccc|c}
                \bottomrule[1pt]
                \makecell[c]{PP logit KD} & \makecell[c]{Label  KD} & TGI & \makecell[c]{Feature KD} & mAPH \\
                \hline
                 & & & & 56.26 \\
                \hline
                $\surd$ &  & & & 57.68 \\
                 &  & $\surd$ &  & 57.61 \\
                \hline
                $\surd$ &  & $\surd$ & & 58.83 \\
                $\surd$ & $\surd$ & & & 58.49 \\
                $\surd$ & $\surd$ & $\surd$ & & \textbf{59.28} \\
                \hline
                $\surd$ & $\surd$ & & $\surd$ & 58.28 \\
                $\surd$ & $\surd$ & $\surd$ & $\surd$ & 58.67 \\
                \toprule[0.8pt]
            \end{tabular}
        }
        \end{small}
        \label{tab:abl_component}
    \end{minipage}
    \hfill
    \begin{minipage}[t]{0.48\hsize}
        \centering
        \caption{Different remap manners studies of TGI based on CP-Voxel-XS and CP-Voxel-XXS.}
        \vspace{-0.1cm}
        \begin{small}
        \setlength\tabcolsep{2pt}
        \scalebox{0.97}{
            \begin{tabular}{c|c|c|c}
                \bottomrule[1pt]
                Student & Teacher & Remap & mAPH \\
                \hline
                \multirow{4}{*}{CP-Voxel-XS} & - & - & 61.16 \\
                 & CP-Voxel & FNA & 62.43 \\
                 & CP-Voxel & OFA & 60.06 \\
                 & CP-Voxel & Slim & 61.74 \\
                \hline
                \multirow{3}{*}{CP-Voxel-XXS} & - & -  & 56.26 \\
                 & CP-Voxel & FNA & 54.92 \\
                 & CP-Voxel-XS & FNA & 57.61 \\
            \toprule[0.8pt]
            \end{tabular}
        }
        \end{small}
        \label{tab:abl_init}
    \end{minipage}
    }}
    \vspace{-0.5cm}
\end{table}

\textbf{Investigation of TGI.} We study how different parameter remapping manners and teachers influence the TGI.
Besides FNA, we also attempt other parameter remapping strategies for TGI, including OFA~\cite{cai2019once} and Slim~\cite{liu2017learning} by selecting important channels with designed indicators. 
As shown in Table~\ref{tab:abl_init}, based on CP-Voxel-XS, OFA and Slim obtain inferior results compared to FNA, since indicator guided channel selection cannot determine consistent channel mapping for skip connection~\cite{he2016deep}, while FNA survives by simply selecting beginning channels of all layers.
In addition, CP-Voxel-XXS only improves by inheriting parameters from CP-Voxel-XS but not CP-Voxel, which indicates a large architecture discrepancy between teacher and student hinders the effectiveness of TGI.


\textbf{Different Variants of PP Logit KD.} Here, we analyze different variants of PP logit KD (\ie pivotal position selection according to teacher confidence, response ranking or soft Gaussian instance mask) in Section~\ref{sec:method} based on CP-Voxel-S. While previous coarse instance-aware response imitation method GID-L obtains around $0.7\%$ improvements, all three PP logit KD variants achieve around $1.6\% \sim 1.9\%$ gains. This demonstrates the significance of focusing on only vital positions in output-level distillation for 3D detection.

\begin{table}[htbp]
    \centering
    \vspace{-0.1cm}
    \makebox[0pt][c]{\parbox{\textwidth}{%
    \begin{minipage}[t]{0.38\hsize}
        \centering
        \caption{Analysis of different PP logit KD variants based on CP-Voxel-S.}
        \vspace{-0.2cm}
        \begin{small}
        \scalebox{0.97}{
            \begin{tabular}{c|c}
                \bottomrule[1pt]
                Logit KD Manner & mAPH \\
                \hline
                None & 62.23 \\
                \hline
                GID-L~\cite{dai2021general} & 62.89 \\
                \hline
                Confidence PP & 63.84 \\
                Rank PP & 63.85 \\
                Gaussian PP & \textbf{64.16} \\
            \toprule[0.8pt]
            \end{tabular}
        }
        \end{small}
        \label{tab:pp_logit_kd}
        \vspace{-0.3cm}
    \end{minipage}
    \hfill
    \begin{minipage}[t]{0.58\hsize}
        \centering
        \caption{Analysis of label KD based on CP-Pillar-v0.48.}
        \vspace{-0.2cm}
        \begin{small}
        \setlength\tabcolsep{3pt}
        \scalebox{0.97}{
            \begin{tabular}{l|cc|cc|c}
                \bottomrule[1pt]
                & \multicolumn{2}{c|}{Ground Truth} &  \multicolumn{2}{c|}{Teacher Pred}  & \multirow{2}{*}{mAPH} \\
                \cline{2-5}
                 & Cls & Reg & Cls & Reg & \\
                \hline
                (a) & all & all & no & no & 56.24 \\
                \hline
                (b) & all & all & all & no & 56.47 \\
                (c) & all & non-overlap & all & all & 56.99 \\
                (d) & all & all & all & non-duplicate & 56.74 \\
                \hline
                (e) & all & all & all & all & \textbf{57.54} \\
            \toprule[0.8pt]
            \end{tabular}
        }
        \end{small}
        \label{tab:label_kd}
        \vspace{-0.3cm}
    \end{minipage}
    }}
\end{table}

\textbf{Analysis of Label KD.} 
Although label KD has been proposed by the recent work~\cite{nguyen2022improving}, it just attributed its improvement to the ``dark knowledge'' from teacher. Here, we attempt to analyze how teacher predictions help student based on CP-Pillar-v0.48, where label KD performs best. As shown in Table~\ref{tab:label_kd}, only integrating teacher prediction for classification label assignment, (b) achieves minor gains, which indicates that label KD mainly benefits the regression objective of student. 
When removing GTs highly overlapped with teacher predictions, (c) still achieves around $0.8\%$ gains, which demonstrates that more achievable regression targets from teacher prediction facilitate student optimization.
Last, by removing teacher predictions with object centers in the same BEV grid (\ie duplicate objectives for single position), we find that (d) suffers performance drop compared to (e), which shows that duplicate regression objectives can also boost student regression ability.




\vspace{-0.3cm}
\section{Conclusion}
\vspace{-0.3cm}
We have examined  the potential of knowledge distillation to serve as a generic method  for obtaining efficient 3D detectors with extensive experimental results and analysis.
We found that pillar-based detector prefers input compression while voxel-based detector is more suitable for width compression in designing efficient student models.
Besides, we proposed pivotal position logit KD and teacher guided initialization for enhancing the 3D KD pipeline.
Our best performing detector outperforms its teacher with $2.4\times$ fewer flops and our most efficient detector is $2.2\times$ faster than previous fastest voxel/pillar-based detector PointPillars on an NVIDIA A100 with higher performance.
We hope our benchmark and analysis could inspire future investigations on this problem.
\vspace{-0.1cm}
\paragraph{Acknowledgement.} This work has been supported by Hong Kong Research
Grant Council - Early Career Scheme (Grant No. 27209621), HKU Startup Fund,
and HKU Seed Fund for Basic Research.

{
\bibliographystyle{plain}
\bibliography{egbib}
}

\section*{Checklist}

\begin{enumerate}

\item For all authors...
\begin{enumerate}
  \item Do the main claims made in the abstract and introduction accurately reflect the paper's contributions and scope?
    \answerYes{}
  \item Did you describe the limitations of your work?
    \answerYes{We do not consider sophisticate layer-wise model compression strategies (see section 3). More details are described in supplemental file.}
  \item Did you discuss any potential negative societal impacts of your work?
    \answerNA{Our work is only for academic research purpose.}
  \item Have you read the ethics review guidelines and ensured that your paper conforms to them?
    \answerYes{}
\end{enumerate}

\item If you are including theoretical results...
\begin{enumerate}
  \item Did you state the full set of assumptions of all theoretical results?
    \answerNA{}
        \item Did you include complete proofs of all theoretical results?
    \answerNA{}
\end{enumerate}

\item If you ran experiments...
\begin{enumerate}
  \item Did you include the code, data, and instructions needed to reproduce the main experimental results (either in the supplemental material or as a URL)?
    \answerYes{Our implementation details are posted in supplemental materials and code will be available.}
  \item Did you specify all the training details (e.g., data splits, hyperparameters, how they were chosen)?
    \answerYes{We follow the popular codebase OpenPCDet (see section 3).}
  \item Did you report error bars (e.g., with respect to the random seed after running experiments multiple times)?
    \answerYes{It is posted in supplemental materials}
  \item Did you include the total amount of compute and the type of resources used (e.g., type of GPUs, internal cluster, or cloud provider)?
    \answerYes{See supplemental materials. Most of experiments are trained with 8 NVIDIA 1080Ti. Few experiments are trained with 8 NVIDIA V100 or NVIDIA A100. Full set results on Waymo are trained with 16 NVIDIA 1080ti.}
\end{enumerate}

\item If you are using existing assets (e.g., code, data, models) or curating/releasing new assets...
\begin{enumerate}
  \item If your work uses existing assets, did you cite the creators?
    \answerYes{See our reference.}
  \item Did you mention the license of the assets?
    \answerNo{They are public released for academic utilization.}
  \item Did you include any new assets either in the supplemental material or as a URL?
    \answerNo{}
  \item Did you discuss whether and how consent was obtained from people whose data you're using/curating?
    \answerYes{They are public released.}
  \item Did you discuss whether the data you are using/curating contains personally identifiable information or offensive content?
    \answerNo{We only use public dataset.}
\end{enumerate}

\item If you used crowdsourcing or conducted research with human subjects...
\begin{enumerate}
  \item Did you include the full text of instructions given to participants and screenshots, if applicable?
    \answerNA{}
  \item Did you describe any potential participant risks, with links to Institutional Review Board (IRB) approvals, if applicable?
    \answerNA{}
  \item Did you include the estimated hourly wage paid to participants and the total amount spent on participant compensation?
    \answerNA{}
\end{enumerate}

\end{enumerate}

\clearpage
\appendix

\section{Appendix}

\centerline{\large{\textbf{Outline}}}
In this supplementary file, we provide more details and experiments not elaborated in our main paper due to page \textcolor{blue}{length} limits:
\begin{itemize}
    \item Sec.~\ref{sec:implmentation_details}: Implementation details and hyper-parameters of our 3D detection KD benchmark.
    \item Sec.~\ref{sec:extra_exper}: Additional experimental results on synergy results of TGI, per-class results, error bar results, and focal loss attempts.
    \item Sec.~\ref{sec:more_analysis}: More analysis, including latency comparisons on different accelerators, operation-level optimizations and detectors, and qualitative analysis of CPR.
    \item Sec.~\ref{sec:supp_generality}: Generality of our method on 3D semantic segmentation.
    \item Sec.~\ref{sec:supp_discussion}: Discussion on other detectors such as sparse detection architectures and other input representations.
    \item Sec.~\ref{sec:limitation}: Limitation analysis.
\end{itemize}

\section{Implementation Details for Our Benchmark}
\label{sec:implmentation_details}
In this section, we describe the implementation of previous 2D KD methods in 3D object detection. Notice that most 2D detection KD methods are built on anchor-based detectors (\eg Faster-RCNN~\cite{ren2015faster}) and model compressed teacher-student pairs, so we modify  them to adapt to anchor-free detectors and handle input resolution compression. On the other hand, we also provide detailed hyper-parameter values to help reproduce our results. Besides, we will also open-source our benchmark suit upon acceptance. Most of the experiments are trained with 8 NVIDIA 1080Ti, while a few experiments are trained with 8 NVIDIA V100 or 8 NVIDIA A100. Full set results on Waymo are trained with 16 NVIDIA 1080Ti or V100.

\textbf{Logit KD.} As for logit KD methods (\ie vanilla KD~\cite{hinton2015distilling} and GID-L~\cite{dai2021general}), vanilla KD use all ones $m_{\text{cls}}$ to fully mimic all teacher outputs and set the mask $m_{\text{cls}}$ to be all one. As for GID-L, the original anchor-wise region selection manner cannot be extended to input resolution compressed students, since the interpolation cannot handle the resolution mismatch of regression predictions $p_\text{reg}$ from teacher and student models. In this regard, we refer to the ablation studies of the original paper and use ground truth boxes as critical region selection criteria. Specifically, we set the non-zero spatial positions in the assigned classification target heatmap to one in $m_{\text{cls}}$.
The loss weight $\alpha_1$ and $\alpha_2$ of $\mathcal{L}^{\text{KD}}_{\text{cls}}$ and $\mathcal{L}^{\text{KD}}_{\text{reg}}$ are set to $15.0$ and $0.2$, respectively.
Notice that the loss weight $\alpha_2$ of regression term $\mathcal{L}^{\text{KD}}_{\text{reg}}$ in Eq. (4) will be set to $0$ for input resolution compressed students.

As for our PP logit KD, the threshold of confidence PP is set to $0.3$ by default and the rank $K$ for rank PP is set to $500$. Although the three variants of PP logit KD show similar performance, we use Gaussian PP by default since it does not need hyper-parameters adjustment when adopted by different teacher-student pairs.

\textbf{Feature KD.} For the implementation of different feature KD methods with Eq. (3), we only employ convolutional block $\phi$ for width compressed students to align the number of channels between $f^s$ and $f^t$ and only hire spatial interpolation $\kappa$ for input compressed students. Besides, for methods that utilize RoI Align $\psi$ to extract object-level features, we will not use the interpolation $\kappa$ to avoid introducing extra interpolation errors.
As for the implementation of each method, we directly align student and teacher full features without mask $m_{\text{feat}}$ and RoI Align $\psi$ in FitNet~\cite{romero2014fitnets}. 
As for Mimic~\cite{li2017mimicking}, we use the more sophisticated RoI Align $\psi$ instead of the original spatial pyramid pooling to extract features for each GT, and construct imitation on the object-level features between teacher and student. 
As for FG~\cite{wang2019distilling}, to extend its anchor-based critical region selection, we directly set the non-zero regions in the assigned classification target heatmap as the critical regions in $m_{\text{feat}}$, the other implementations are the same as FitNet. 
As for GID-F~\cite{dai2021general}, we use teacher predictions after Non-Maximum Suppression (NMS) as critical regions and set their corresponding spatial positions in $m_{\text{feat}}$ to one. Besides, we utilize RoI Align $\psi$ to extract object-level features to calculate feature distillation loss and also apply the relation loss among different object features as the description in the original paper. 
The loss weight $\alpha_3$ of feature KD loss $\mathcal{L}^{\text{KD}}_{\text{feat}}$ is set to $100$ for input compressed students or $200$ for width compressed students, respectively. The loss weight of the relation loss of GID-F is set to $0.1$.

\textbf{Label KD.} There is only one work for label KD which has been described in the main paper. Notice that we do not hire NMS for teacher predictions for label assignment empirically. 
The score threshold $\tau$ to filter high-quality teacher predictions is set to $0.6$ by default.

Among different 2D KD methods, we notice that FG~\cite{wang2019distilling}, Mimic~\cite{li2017mimicking} and GID~\cite{dai2021general} highlight the critical region selection on feature KD or logit KD to tackle the imbalance between foreground and background regions in object detection. There are mainly two foreground-region imitation strategies: one is using RoI Align $\psi$ to extract object-wise features with teacher prediction or GT boxes as guidance (\eg Mimic~\cite{li2017mimicking} and GID-F~\cite{dai2021general}); the other is assigning a one-hot mask $m_{\text{feat}}$ or $m_{\text{cls}}$ to calculate imitation loss on some critical regions (\eg FG~\cite{wang2019distilling} and GID-L~\cite{dai2021general}). 
We empirically find that RoI Align is more suitable for input compression setups since it avoids the interpolation errors when aligning spatial resolutions while the mask-based strategy is more flexible and general as it allows position-wise imitation. 
Comparing different critical region selection techniques, we empirically show that a key factor to make KD methods work well on 3D detection is to focus on only a few positions. For example, our proposed PP logit KD focus on only around $\frac{1}{5} \sim \frac{1}{20}$ positions of the positions selected by traditional 2D KD methods. This is caused by the fact that a spatial position in the 3D BEV features can represent a $0.8m \times 0.8m \times 6m$ pillar in the 3D geometry space, which is informative and can cover even a single pedestrian. In this regard, the critical region selection techniques are supposed to focus on fewer informative positions in the 3D detection setting.

\section{Additional Experimental Results}
\label{sec:extra_exper}
In this section, we provide some additional experimental results as a supplement to our main paper. This part consists of the full synergy results of TGI on six teacher-student pairs, per-class performance and error bar results.

\subsection{Synergy Results of TGI}
\begin{table}[htbp]
    \centering
    \vspace{-0.2cm}
    \caption{Synergy results of TGI and feature KD on Waymo with six teacher-student pairs. Performance are measured in LEVEL 2 mAPH. Teacher results are masked by gray. }
    \vspace{-0.2cm}
    \begin{small}
    \setlength\tabcolsep{3pt}
    \scalebox{0.93}{
        \begin{tabular}{l|c||c|c|c|c|c|c|c|c}
            \bottomrule[1pt]
            \multirow{2}{*}{Detector} & \multirow{2}{*}{\makecell[c]{No \\ Distill}} & \multirow{2}{*}{\makecell[c]{Feature \\ KD}} & \multirow{2}{*}{\makecell[c]{Label \\ KD}} & \multirow{2}{*}{\makecell[c]{PP Logit \\ KD}} & \multirow{2}{*}{TGI} & \multirow{2}{*}{\makecell[c]{PP Logit KD \\ + Feature KD}} & \multirow{2}{*}{\makecell[c]{PP Logit KD  \\ + TGI}}  & \multirow{2}{*}{\makecell[c]{Label KD \\ + Feature KD}} & \multirow{2}{*}{\makecell[c]{Label KD \\+ TGI}} \\
            & & & & & & & & & \\
            \hline
            \rowcolor{Gray!16} CP-Pillar & 59.09 & - & - & - & - & - & - & - & - \\
            CP-Pillar-v0.4 & 57.55 & 58.57 & 58.10 & 58.21 & 59.03 &  58.18 & 59.24 & 58.35 & 59.19 \\
            CP-Pillar-v0.48 & 56.27 & 57.26 & 57.54 & 56.89 & 57.91 & 57.11 & 58.20 & 57.43 & 58.34 \\
            CP-Pillar-v0.64 & 52.81 & 53.83 & 53.78 & 54.32 & 54.30 & 54.14 & 55.55 & 54.24 & 55.59 \\
            \toprule[1pt]
            \bottomrule[1pt]
            \rowcolor{Gray!16} CP-Voxel & 64.29 & - & - & - & - & - & - & - & - \\
            CP-Voxel-S & 62.23 & 63.35 & 63.31 & 64.16 & 63.48 & 63.58 &  64.18 & 62.62 & 63.50 \\
            CP-Voxel-XS & 61.16 & 62.48 & 61.81 & 62.76 & 62.43 & 62.90 & 63.41 & 62.34 & 62.85 \\
            CP-Voxel-XXS & 56.26 & 57.92 & 57.02 & 57.68 & 57.61 & 58.19 & 58.83 & 57.06 & 57.71 \\
            \toprule[0.8pt]
        \end{tabular}
    }
    \end{small}
    \label{tab:TGI_all_synergy}
    \vspace{-0.3cm}
\end{table}

The poor synergy effect of feature KD is the main motivation for us to design TGI. Due to the page limitation, we only present its experimental results on CP-Voxel-XXS as an example. Here, we compare the synergy results of feature KD and TGI on six teacher-student pairs to further show the promising performance of our TGI. As shown in Table~\ref{tab:TGI_all_synergy}, the results obtained by combining TGI and label KD or PP logit KD consistently outperform the synergy results of feature KD on all 12 scenarios, manifesting that TGI collaborates better with other KD techniques. Furthermore, our TGI itself achieves comparable results or even surpasses feature KD among six teacher-student pairs. These experimental results strongly demonstrate that our proposed TGI can be a powerful substitute for feature KD to transfer the feature extraction ability from the teacher model.

\subsection{Per-class Performance}
\begin{table}[htbp]
    \centering
    \vspace{-0.2cm}
    \caption{Per-class performance on full Waymo dataset for our six distilled efficient student models. Performance are measured in mAP/mAPH. Teacher results are masked by gray. Best results are indicated by \textbf{bold}.}
    \vspace{-0.2cm}
    \begin{small}
    \setlength\tabcolsep{3.9pt}
    \scalebox{1.0}{
        \begin{tabular}{l|c|c|c|c|c|c}
            \bottomrule[1pt]
            \multirow{2}{*}{Detector} & \multicolumn{2}{c|}{Vehicle} & \multicolumn{2}{c|}{Pedestrian} & \multicolumn{2}{c}{Cyclist} \\
            \cline{2-7}
            & LEVEL 1 & LEVEL 2 & LEVEL 1 & LEVEL 2 & LEVEL 1 & LEVEL 2 \\
            \hline
            \rowcolor{Gray!16} CP-Pillar & 72.75/72.24 & 64.48/64.02 & 74.01/64.06 & 65.74/56.76 & \textbf{67.84}/\textbf{66.37} & \textbf{65.34}/\textbf{63.92} \\
            CP-Pillar-v0.4 + Ours & \textbf{73.01}/\textbf{72.46} & \textbf{64.85}/\textbf{64.36} & \textbf{75.00}/\textbf{64.24} & \textbf{66.86}/\textbf{57.13} & 67.18/65.76 & 64.69/63.32 \\
            CP-Pillar-v0.48 + Ours & 72.42/71.85 & 64.42/63.89 & 74.38/63.74 & 66.26/56.62 & 66.19/64.76 & 63.72/62.34 \\
            CP-Pillar-v0.64 + Ours & 71.37/70.77 & 63.30/62.75 & 71.45/61.05 & 63.22/53.86 & 63.87/62.39 & 61.48/60.05 \\
            \toprule[1pt]
            \bottomrule[1pt]
            \rowcolor{Gray!16} CP-Voxel & \textbf{74.31}/\textbf{73.75} & \textbf{66.35}/\textbf{65.84} & 76.19/70.10 & 68.44/62.82 &  71.76/70.63 & 69.16/68.07 \\
            CP-Voxel-S + Ours & 74.28/73.72 & 66.17/65.66 & \textbf{76.72}/\textbf{70.68} & \textbf{68.96}/\textbf{63.37} & \textbf{71.97}/\textbf{70.81} & \textbf{69.36}/\textbf{68.24} \\
            CP-Voxel-XS + Ours & 73.62/73.05 & 65.53/65.01 & 75.50/69.29 & 67.67/61.96 & 71.30/70.09 & 68.69/67.52 \\
            CP-Voxel-XXS + Ours & 69.20/68.55 & 61.15/60.57 & 71.76/64.95 & 63.71/57.53 & 68.51/67.18 & 65.98/64.70 \\
            \toprule[0.8pt]
        \end{tabular}
    }
    \end{small}
    \label{tab:per_class}
\end{table}

We report the per-category performance of our efficient detectors on full Waymo Open Dataset~\cite{sun2020scalability} in Table~\ref{tab:per_class}. As illustrated in Table~\ref{tab:per_class}, comparing CP-Pillar-v0.64 and CP-Pillar, the performance gap mainly lies in pedestrians and cyclists (around $3\%$ gap), while vehicle suffers around $1.5\%$ gap. This might be caused by the fact that coarser input resolution penalizes the performance of small objects such as pedestrians and cyclists more severely. As for voxel-based detector CP-Voxel-XXS, we notice that its performance gap from teacher distributes more evenly on different categories than CP-Pillar-v0.64, as the model width compression does not have special penalization on any categories.

\subsection{Error Bar}
\begin{table}[htbp]
    \centering
    \caption{Repeat results of our different models on Waymo . We report the reproduced results with 5 rounds as well as their averaged results and standard variance. The performance is measured in LEVEL 2 mAPH. }
    \vspace{-0.2cm}
    \begin{small}
    \setlength\tabcolsep{3.8pt}
    \scalebox{1.0}{
        \begin{tabular}{l|c|c|c|c|c|c|c}
            \bottomrule[1pt]
            Detector & Round 1 & Round 2 & Round 3 & Round 4 & Round 5 & Average & Standard Variance \\
            \hline
            \rowcolor{Gray!16} CP-Pillar & 59.09 & 59.13 & 59.13 & 59.14 & 59.01 & 59.10 & 0.05 \\
            \hline
            CP-Pillar-v0.64 & 52.81 & 52.75 & 52.85 & 52.85 & 53.25 & 52.90 & 0.20 \\
            \hline
            CP-Pillar-v0.64 + Ours & 55.75 & 55.82 & 56.02 & 55.75 & 55.73 & 55.81 & 0.12 \\
            \toprule[0.8pt]
        \end{tabular}
    }
    \end{small}
    \label{tab:error_bar}
    \vspace{-0.3cm}
\end{table}

Here, to show the robustness of our experimental results, we reproduce knowledge distillation on CP-Pillar-v0.64 five times and report the average and standard derivation of performance. As shown in Table~\ref{tab:error_bar}, the performance of our distilled CP-Pillar-v0.64 is more stable than the student without distillation, which indicates that our improved KD pipeline can boost performance stably.

\subsection{Focal Loss Results}
As focal loss~\cite{lin2017focal} is a widely-used solution for the foreground and background region imbalance issue, it is intuitive to also employ it as the distillation loss. In this regard, here we provide an experimental comparison between focal loss and our proposed PP logit KD for logit KD. As shown in Table~\ref{tab:focal_loss}, PP logit KD is around 0.7\% and 8.2\% higher than focal loss on CP-Voxel-XS and CP-Pillar-v0.64, respectively. As for CP-Pillar-v0.64, since the capability difference between teacher and student are large, focal loss even suffers performance degradation compared to vanilla KD, while our PP logit KD consistently brings performance boost. The reason for the inferior performance of focal loss for distillation is that it will emphasize regions that are most different among teacher and student pairs but not most information-rich areas. Those large prediction difference areas could be caused by the capability gap between teacher and student and thus renders focal loss a suboptimal strategy for student learning.

\begin{table}[htbp]
    \centering
    \caption{Results of leveraging focal loss as logit distillation loss on Waymo. Teacher models are marked in gray.}
    \vspace{-0.2cm}
    \begin{small}
    \scalebox{1.0}{
        \begin{tabular}{l|c|c|c|c}
            \bottomrule[1pt]
            Detector & No Distill & KD~\cite{hinton2015distilling} & Focal loss & PP Logit KD \\
            \hline
            \rowcolor{Gray!16} CP-Voxel & 64.29 & - & - & - \\
            \hline
            CP-Voxel-XS & 62.23 & 62.81 & 63.48 & 64.16 \\
            \toprule[1pt]
            \bottomrule[1pt]
            \rowcolor{Gray!16} CP-Pillar & 59.09 & - & - & - \\
            \hline
            CP-Pillar-v0.64 & 52.81 & 50.78 & 46.11 & 54.32 \\
            \toprule[0.8pt]
        \end{tabular}
    }
    \end{small}
    \label{tab:focal_loss}
    \vspace{-0.3cm}
\end{table}

\section{More Analysis}
\label{sec:more_analysis}
In this section, we provide some investigations on the influence of accelerator types and operation-level optimizations on the measured latency as well as the qualitative analysis of our proposed CPR.

\subsection{Latency Analysis}
\begin{figure}[h]
    \centering
    \vspace{-0.3cm}
    \includegraphics[width=1\linewidth]{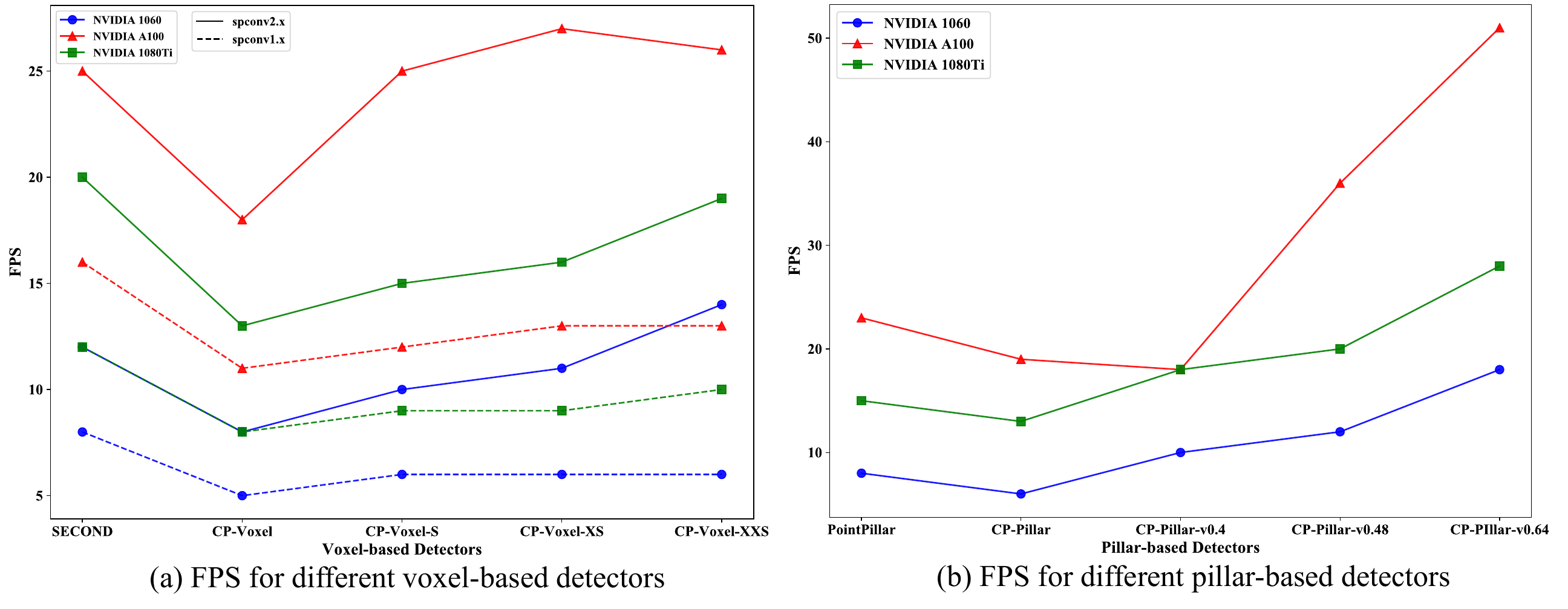}
    \vspace{-0.8cm}
    \caption{Comparison on the FPS with different hardware devices and operation-level optimizations for different detectors.}
    \label{fig:latency}
\end{figure}

\begin{table}[htbp]
    \centering
    \vspace{-0.1cm}
    \caption{Information of the inference machine. }
    \vspace{-0.2cm}
    \begin{small}
    \setlength\tabcolsep{3.8pt}
    \scalebox{1.0}{
        \begin{tabular}{l|c|c}
            \bottomrule[1pt]
            Type & GPU & CPU \\
            \hline
            Personal Computer & NVIDIA GTX-1060 & Intel(R) Core(TM) i7-7700K CPU @ 4.20GHz  \\
            \hline
            Server\#1 & NVIDIA GTX-1080Ti & Intel(R) Xeon(R) CPU E5-2682 v4 @ 2.50GHz \\
            \hline
            Server\#2 & NVIDIA A100 & AMD EPYC 7742 64-Core \\
            \toprule[0.8pt]
        \end{tabular}
    }
    \end{small}
    \label{tab:machine_info}
    \vspace{-0.3cm}
\end{table}

Inference time, latency or FPS directly measure the execution speed of a model on a given hardware configuration, and have been widely adopted to assess the model efficiency in 3D detection. However, different papers~\cite{zhang2022not,yin2021center,deng2020voxel} measure the latency based on different machines, hindering the fair comparisons and standardization among different approaches. Besides, we empirically show that the operation-level optimization has a large impact on the latency measurement and even influences the conclusion. In this regard, we investigate how different hardware devices and operation-level optimizations (\ie sparse convolution~\cite{graham2017submanifold}) affect the latency measurement in terms of FPS. The basic information of our tested machines is shown in Table~\ref{tab:machine_info}, including one personal computer and two servers.

\textbf{Influence of Hardware Devices.} As shown in Figure~\ref{fig:latency} (a) and (b), detectors run faster on more powerful GPU consistently for both voxel-based and pillar-based detectors, which demonstrate the great influence of hardware devices. In addition, we notice that some compressed detectors meet negative optimization (\ie model needing fewer computations has larger latency) on the latest GPU architecture NVIDIA-A100 (see CP-Voxel-XS \vs CP-Voxel-XXS and CP-Pillar \vs CP-Pillar-v0.4), which has not been observed on GTX-1060 and GTX-1080Ti. This might be caused by the different underlying implementation strategies at the hardware level. This also verifies that latency-orientated efficient detector designs largely depend on the type of hardware devices.

\textbf{Operation-level Optimization.} Sparse convolution network~\cite{graham2017submanifold} is a major component in existing voxel-based detectors to efficiently extract voxel-wise features from voxelized point clouds. 
As they are not fully-optimized toward hardware, it occupies a large percentage of latency for voxel-based detectors using A popular implementation Spconv~\footnote{https://github.com/traveller59/spconv}, although the GLOPs are not that high. Here, we also investigate how different implementations of the sparse convolution influence the measured latency.
As shown in Figure~\ref{fig:latency} (a), voxel-based detectors implemented with Spconv2.x is much faster than their counterparts with Spconv1.x (Spconv2.x is the optimized version of Spconv1.x). Moreover, the operation-level optimization can even have a larger impact than the hardware devices (see NVIDIA 1080Ti with Spconv1.x and NVIDIA 1060 with Spconv2.x). In addition, although the width-level compressed students of CP-Voxel require significantly fewer flops, parameters and activations compared to CP-Voxel, they cannot obtain obvious speed up on the latency with Spconv1.x, which indicates that non-parametric computations (\ie computation not directly related to learnable parameters) occupy most of the latency for voxel-based detectors using Spconv1.x. As a consequence, based on NVIDIA 1060, CP-Voxel with Spconv2.x runs faster than CP-Pillar, while runs slower than CP-Pillar when using Spconv 1.x. This demonstrates that testing model latency on different operation-level optimizations can draw totally different conclusions when comparing the efficiency of different detectors in terms of latency.

Besides the above two factors, we also observe that even hardware status (\eg temperature) could influence the final obtained latency, indicating that the latency is hard to stably reproduce on the same machine and cannot serve as a standard measurement on model efficiency. In this regard, we focus on the parametric measurement such as flops and activations in the main paper, as they will not be influenced by the above hardware, software or environment level factors.

\subsection{Qualitative Analysis of CPR}
\begin{table}[htbp]
    \centering
    \vspace{-0.3cm}
    \caption{More model compression results. Teacher models are marked in gray. See text for details.}
    \vspace{-0.3cm}
    \begin{small}
    \setlength\tabcolsep{3pt}
    \scalebox{0.97}{
        \begin{tabular}{cc|c|c|c|c|c|c|c|c|c|c|c|c}
            \bottomrule[1pt]
            \multicolumn{7}{c|}{Architecture} & \multicolumn{5}{c|}{Efficiency} & \multirow{3}{*}{\makecell[c]{LEVEL 2 \\ mAPH}} & \multirow{3}{*}{CPR} \\
            \cline{1-12}
            \multicolumn{2}{c|}{\multirow{2}{*}{Detector}} & \multicolumn{3}{c|}{Width} & \multicolumn{2}{c|}{Depth} &  Params  & Flops & Acts & Latency & Mem. & & \\
            \cline{3-7}
            & & PFE & BFE & Head & PFE & BFE & (M) & (G) & (M) & (ms) & (G) & &  \\
            \hline
            \cellcolor{white!25} \multirow{5}{*}{CP-Pillar} &  & \cellcolor{Gray!16} 1.00 & \cellcolor{Gray!16} 1.00 & \cellcolor{Gray!16} 1.00 & \cellcolor{Gray!16} 1.00 & \cellcolor{Gray!16} 1.00 & \cellcolor{Gray!16} 5.2 & \cellcolor{Gray!16} 333.9 & \cellcolor{Gray!16} 303.0 & \cellcolor{Gray!16} 157.9 & \cellcolor{Gray!16} 5.2 & \cellcolor{Gray!16} 59.09 & \cellcolor{Gray!16} - \\
            \cline{3-14}
            & (a) & \cellcolor{blue!10} 1.00 & \cellcolor{blue!10} 0.50 & \cellcolor{blue!10} 1.00 & 1.00 & 1.00 & 1.5 & 130.1 & 203.1 & 97.1 & 3.6 & 55.35 & 0.58 \\
            \cline{3-14}
            & (b) & \cellcolor{blue!10} 1.00 & \cellcolor{blue!10} 0.25 & \cellcolor{blue!10} 0.25 & 1.00 & 1.00 & 0.3 & 23.8 & 91.2 & 51.9 & 2.3 & 46.16 & 0.59 \\
            \cline{3-14}
            & (c) & 1.00 & 1.00 & 1.00 & \cellcolor{blue!10} 1.00 & \cellcolor{blue!10} 0.50 & 2.2 & 258.5 & 234.1 & 118.5 & 4.3 & 55.24 & 0.52 \\
            \cline{3-14}
            & (d) & 1.00 & 1.00 & 1.00 & \cellcolor{blue!10} 1.00 & \cellcolor{blue!10} 0.33  & 1.4 & 234.6 & 210.0 & 107.8 & 4.0 & 47.97 & 0.42 \\
            \toprule[1pt]
            \bottomrule[1pt]
            \cellcolor{white!25} \multirow{5}{*}{CP-Voxel} & & \cellcolor{Gray!16} 1.00 & \cellcolor{Gray!16} 1.00 & \cellcolor{Gray!16} 1.00 & \cellcolor{Gray!16} 1.00 & \cellcolor{Gray!16} 1.00 & \cellcolor{Gray!16} 7.8 & \cellcolor{Gray!16} 114.7 & \cellcolor{Gray!16} 101.9 & \cellcolor{Gray!16} 125.7 & \cellcolor{Gray!16} 2.8 & \cellcolor{Gray!16} 64.29 & \cellcolor{Gray!16}- \\
            \cline{3-14}
            & (a) & \cellcolor{blue!10} 0.50 & \cellcolor{blue!10} 0.50 & \cellcolor{blue!10} 0.50 & 1.00 & 1.00 & 1.9 & 28.8 & 51.2 & 75.1 & 1.7 & 59.47 & 0.64 \\
            \cline{3-14}
            & (b) & \cellcolor{blue!10} 0.50 & \cellcolor{blue!10} 0.25 & \cellcolor{blue!10} 0.25 & 1.00 & 1.00 & 1.0 & 12.0 & 33.1 & 70.4 & 1.3 & 56.26 & 0.67 \\
            \cline{3-14}
            & (c) & 1.00 & 1.00 & 1.00 & \cellcolor{blue!10} 0.50 & \cellcolor{blue!10} 0.50 & 3.0 & 63.9 & 65.2 & 73.0 & 1.9 & 60.95 & 0.61 \\
            \cline{3-14}
            & (d) & 1.00 & 1.00 & 1.00 & \cellcolor{blue!10} 0.33 & \cellcolor{blue!10} 0.33 & 1.8 & 47.9 & 52.2 & 59.0 & 1.6 & 55.78 & 0.57 \\
            \toprule[0.8pt]
        \end{tabular}
    }
    \end{small}
    \label{tab:CPR}
    \vspace{-0.3cm}
\end{table}

When designing efficient student models, we propose CPR to quantitatively measure the trade offs between efficiency and performance of a compressed student model. Here, we take some examples to qualitatively analyze the correlation between the CPR, the model efficiency as well as the model accuracy. As shown in Table~\ref{tab:CPR}, comparing CP-Pillar (a) and (c), they achieve similar performance, but CP-Pillar (a) requires only half of flops and fewer parameters, activations, latency and training memory. This indicates that CP-Pillar (a) achieves better trade offs between accuracy and efficiency, which can be reflected on its higher CPR. Similar conclusions can be drawn by comparing CP-Pillar (b) and (d), CP-Voxel (a) and (c), as well as CP-Voxel (b) and (d). These qualitative results demonstrate the good correlation between CPR and the compromise between accuracy and efficiency for a given compressed model.

\section{Generality on 3D Semantic Segmentation}
\label{sec:supp_generality}
In this work, the above experiments are all built on 3D object detection, which is a sparse prediction task. However, we argue that our sparse distillation manner (\ie pivotal position logit KD) can also generalize to dense prediction tasks such as 3D semantic segmentation. As the student model has dense GTs supervision in training, dense distillation loss on massive uninformative points and regions, such as road points, might be redundant and can overwhelm the overall distillation loss. Instead, our sparse distillation might help the student focus on more important areas by using teacher prediction as regularization.

Here, we follow the design principle of PP logit KD and adapt it to handle the dense semantic segmentation task. We apply distillation loss on points with predictions that are correct but less confident than the teacher. Our simple design is motivated by three intuitions: ($i$) Points that are correctly predicted with lower confidence are often some challenging cases that the model is struggling but also has the capability to handle. By harvesting knowledge from a high-performing teacher model, the student can learn to match the confidence level of the teacher which provides more information than the one-hot GT. ($ii$) Points that are correctly predicted with higher confidence are often easy samples that have very close prediction confidence to the teacher model. Considering that these samples are already handled well by the model, they have low chance to benefit from distillation but might cause redundancies. ($iii$) Points that are incorrectly predicted by the student are often cases that might be out of the ability of student models. Specifically, we have the confidence of student predictions $\text{conf}^s$, the confidence of teacher predictions $\text{conf}^t$ and a pre-defined threshold $\tau$. We will only apply distillation loss for student predictions that are correct and have $\text{conf}^s + \tau < \text{conf}^t$.

We also provide experimental results for our design on the 3D semantic segmentation dataset ScanNet~\cite{dai2017scannet}. Here, we use a small version of MinkowskiNet~\cite{choy20194d} for fast verification. On the one hand, as shown in Table~\ref{tab:semseg}, we try both model width and input resolution compression to obtain efficient student models, and select MinkowskiNet14-v0.04 as the student model for KD due to its higher CPR. On the other hand, as shown in Table~\ref{tab:semseg_kd}, we compare the effectiveness of KD~\cite{hinton2015distilling}, PP logit KD and TGI on MinkowskiNet14-v0.04, where both our proposed PP logit KD and TGI obtain improvements. 
In particular, our sparse PP logit KD surpasses the dense logit KD method with around 0.8\% gains. Our statistics also show that our PP logit KD only leverages 19.03\% points for distillation at the first epoch and 3.66\% points for distillation at the last epoch. These experiments and statistics demonstrate that sparse distillation can also work on the dense prediction task.

\begin{table}[htbp]
    \centering
    \caption{Model width and input resolution compression results of MinkowskiNet14 on ScanNet. The teacher model is marked in gray.}
    \vspace{-0.3cm}
    \begin{small}
    \scalebox{1.0}{
        \begin{tabular}{c|c|c|c|c|c|c|c}
            \bottomrule[1pt]
            \multicolumn{3}{c|}{Architecture} & \multicolumn{3}{c|}{Efficiency} & \multirow{2}{*}{mIoU} & \multirow{2}{*}{CPR} \\
            \cline{1-6}
            Model & {Width} & Voxel Size (m) &  Params (M) & Flops (T) & Acts (M) & & \\
            \hline
            \rowcolor{Gray!16} MinkowskiNet14 & 1.0 & 0.02 & 1.7 & 46.2 & 27.9 & 65.77 & - \\
            \hline
            MinkowskiNet14-w0.5 & \cellcolor{blue!10}0.5 & 0.02 & 0.5 & 18.2 & 17.4 & 61.84 & 0.60 \\
            \hline
            MinkowskiNet14-v0.04 & 1.0 & \cellcolor{blue!10}0.04 & 1.7 & 5.7 & 8.9 & 62.82 & 0.78 \\
            \toprule[0.8pt]
        \end{tabular}
    }
    \end{small}
    \label{tab:semseg}
\end{table}

\begin{table}[htbp]
    \centering
    \caption{Knowledge distillation results of compressed MinkowskiNet14 on ScanNet.}
    \vspace{-0.3cm}
    \begin{small}
    \scalebox{1.0}{
        \begin{tabular}{c|c|c|c|c|c|c|c}
            \bottomrule[1pt]
            Model & Role & No Distill &  KD~\cite{hinton2015distilling} & PP Logit KD & TGI & Flops (T) & Acts (M) \\
            \hline
            MinkowskiNet14 & Teacher & 65.77 & - & - & - & 46.2 & 27.9 \\
            \hline
            MinkowskiNet14-v0.04 & Student & 62.82 & 63.65 & 64.40 & 64.22 & 5.7 & 8.9 \\
            \toprule[0.8pt]
        \end{tabular}
    }
    \end{small}
    \label{tab:semseg_kd}
    \vspace{-0.3cm}
\end{table}

\section{Discussion on Other Detectors}
\label{sec:other_detectors}
In this work, we mainly focus on dense detectors (\eg CenterPoint~\cite{yin2021center}, PV-RCNN++~\cite{shi2021pv}, SECOND~\cite{yan2018second}) with the most popular pillar/voxel input representations. 
Here, we construct some discussion for the knowledge distillation on the sparse detectors and other input representations to further demonstrate the generality of our distillation manners.

\label{sec:supp_discussion}
\subsection{Discussion for Sparse Detectors}
As the emergency of DETR~\cite{carion2020end}, object detectors that directly produce sparse prediction without post-processing become a new popular detection paradigm. 
Although we only try our KD manner on dense detector in the main paper, we argue that our sparse distillation is still applicable for sparse transformer-based detector such as DETR~\cite{carion2020end}, Deformable DETR~\cite{zhu2020deformable}, Object DGCNN~\cite{wang2021object}, etc.

\begin{figure}[h]
    \centering
    \vspace{-0.1cm}
    \includegraphics[width=1\linewidth]{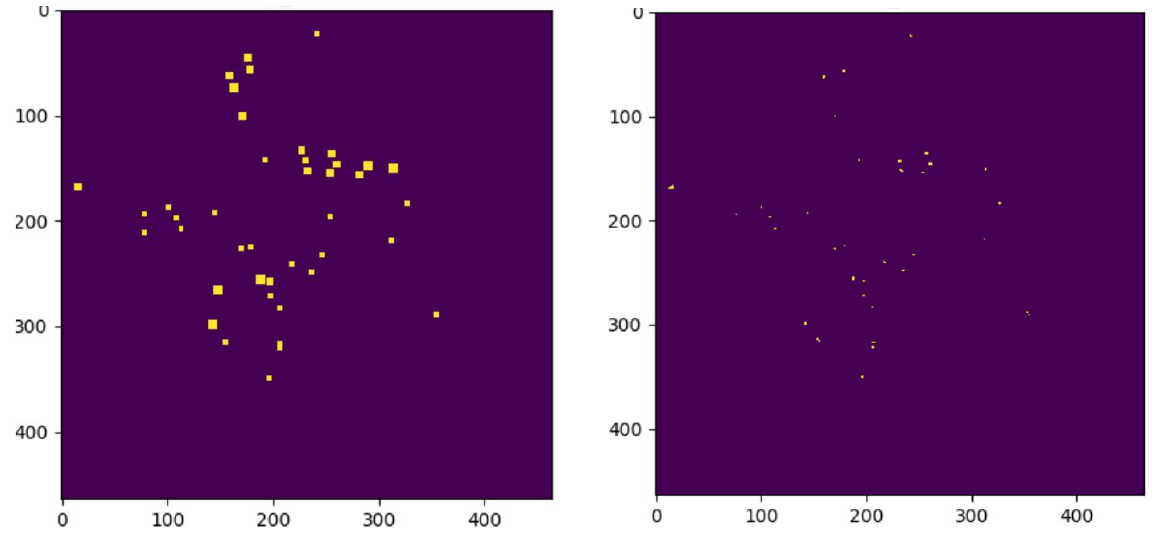}
    \vspace{-0.5cm}
    \caption{Visualization comparison of the imitation position (yellow positions) between instance-wise KD and our PP KD on the bird's eye view. Left: valid imitation regions for instance-wise KD. Right: valid imitation positions for our PP KD. Our PP KD has more fine-grained imitation regions compared to instance-wise KD. Best viewed in color.}
    \label{fig:kd_vis_compare}
\end{figure}

On the one hand, sparse transformer-based detectors that directly make instance predictions actually rely on learning to some sparser reference points and corresponding position features. For example, each object query in Deformable DETR~\cite{zhu2020deformable} or Object DGCNN~\cite{wang2021object} is decoded into a reference point and  neighboring points in order to focus only on those most informative positions. On the other hand, although sparse detectors can directly generate sparse instance predictions, our sparse distillation (\ie pivotal position KD) focuses on sparser and more fine-grained position-level information (see Figure~\ref{fig:kd_vis_compare}). In this regard, it should still be applicable to sparse models with some specific modifications.

Here, we take Object DGCNN~\cite{wang2021object} as an example and provide two possible sparse distillation designs.

(1) As the transformer encoder and decoder of Object DGCNN are similar to Deformable DETR, it can be simply extended to a two-stage variant as Deformable DETR. In the two-stage variant, the transformer encoder will regard each pixel as an object query and construct a dense scoring on it, where top-score positions are picked as reference points. This is similar to our designed rank PP KD which enforces the student to imitate the prediction of teacher top-rank positions. Therefore, we can directly apply our sparse rank PP KD to those dense scoring predictions between teacher and student. Besides, we will also carry on feature imitation on those teacher top-ranked positions between teacher and student.

(2) As for the one-stage variant of sparse detectors, learnable object queries will be decoded into reference points and neighboring points, so the sparse distillation can be constructed on those points and their corresponding BEV features. Specifically, we can first match the positive object queries of teacher and student as query pairs by checking whether they are matched to the same GT box. Then, we can enforce the decoded reference and neighboring points of the student to mimic their paired teacher counterparts. Besides, we will construct imitation on BEV features of those reference and neighboring positions between teacher and student.

\subsection{Discussion for other input representations}
Apart from the most popular pillar/voxel based object detectors discussed in the main paper, there are also point-based and range image based detectors. Therefore, we also provide the discussion on the point-based and range-based detectors here.
As the TGI and label KD are detector-agnostic distillation manners and can be easily extended to detectors with any input representations, we only discuss the sparse distillation -- pivotal position KD here. 


Point-based detector~\cite{shi2019pointrcnn,chen2019fast,yang2019std,zhang2022not} take raw point clouds as input and and employed PointNet++~\cite{qi2017pointnet++} to extract point features and generate point-wise object proposals.
Range image based detectors leverage the native and dense representation for 3D points captured from LiDAR~\cite{meyer2019lasernet,liang2020rangercnn,bewley2020range,sun2021rsn}.
As for the knowledge distillation on point-based and range image based detectors, since they still need to generate dense point-wise object proposals, our sparse distillation can still directly apply to it by selecting confident or top-ranked teacher positions for imitation (\ie Confidence PP and Rank PP in Table~\ref{tab:pp_logit_kd}). In this regard, our distillation strategies should be generalizable to all existing input representations.

\section{Limitations}
\label{sec:limitation}
Although our work has already investigated the compression on model width, model depth and input resolution for designing lightweight student detectors, there are also exhausted layer-wise model compression methods~\cite{liu2017learning,li2020eagleeye}, which have not been attempted in this work. 
Although missing the compression attempts in such perspective, we argue that our paper mainly focuses on exploring the potential of knowledge distillation to obtain efficient 3D detectors and the existing compression attempts already fulfill our demands. Besides, we believe that the further progress and investigation of designing efficient 3D detectors is orthogonal to our KD attempts and can cooperate with our improved KD pipeline to obtain more efficient detectors.


\end{document}